\documentclass{article} 

\usepackage{paper-style,times}

\usepackage{amsmath,amsfonts,bm}

\def\eqref#1{equation~\ref{#1}}

\def\1{\bm{1}}

\DeclareMathAlphabet{\mathsfit}{\encodingdefault}{\sfdefault}{m}{sl}
\SetMathAlphabet{\mathsfit}{bold}{\encodingdefault}{\sfdefault}{bx}{n}

\usepackage{fontawesome5}
\usepackage{hyperref}
\usepackage{url}
\usepackage{graphicx}
\usepackage{float}
\usepackage{xspace}
\usepackage{makecell}
\usepackage{array}
\usepackage{colortbl}
\usepackage{multirow}
\usepackage[normalem]{ulem}
\usepackage{subcaption}
\usepackage{tabularx}
\usepackage{color}
\usepackage{pifont}
\usepackage[utf8]{inputenc}
\usepackage[T1]{fontenc}
\usepackage{CJKutf8}
\usepackage{enumitem}
\usepackage{diagbox}
\usepackage{listings}
\usepackage[flushleft]{threeparttable}
\usepackage{booktabs}
\usepackage{xcolor}      
\usepackage{caption}     
\usepackage{booktabs}    
\usepackage{colortbl}    
\usepackage{wrapfig}
\usepackage{afterpage}
\usepackage{algorithmicx,algorithm}
\usepackage{threeparttable}
\usepackage{xcolor}
\definecolor{redorange}{rgb}{1, 0.5, 0}
\definecolor{darkgreen}{rgb}{0.0, 0.5, 0.0}
\definecolor{darkred}{rgb}{0.7, 0.0, 0.0}
\usepackage{pifont}
\usepackage{tikz}
\usepackage{bm}

\usepackage{marvosym}

\renewcommand{\thefootnote}{\raisebox{0.5pt}{\scriptsize \faEnvelope}}

\renewcommand{\thefootnote}{\fnsymbol{footnote}}
\makeatletter
\def\@fnsymbol#1{%
  \ifcase#1\or \faEnvelope \or \dagger \or \ddagger \or
  \mathsection \or \mathparagraph \or \| \or ** \or \dagger\dagger
  \or \ddagger\ddagger \else \@ctrerr \fi}
\makeatother

\newcommand{\icon}{\raisebox{-4.9pt}{\includegraphics[width=1.1em]{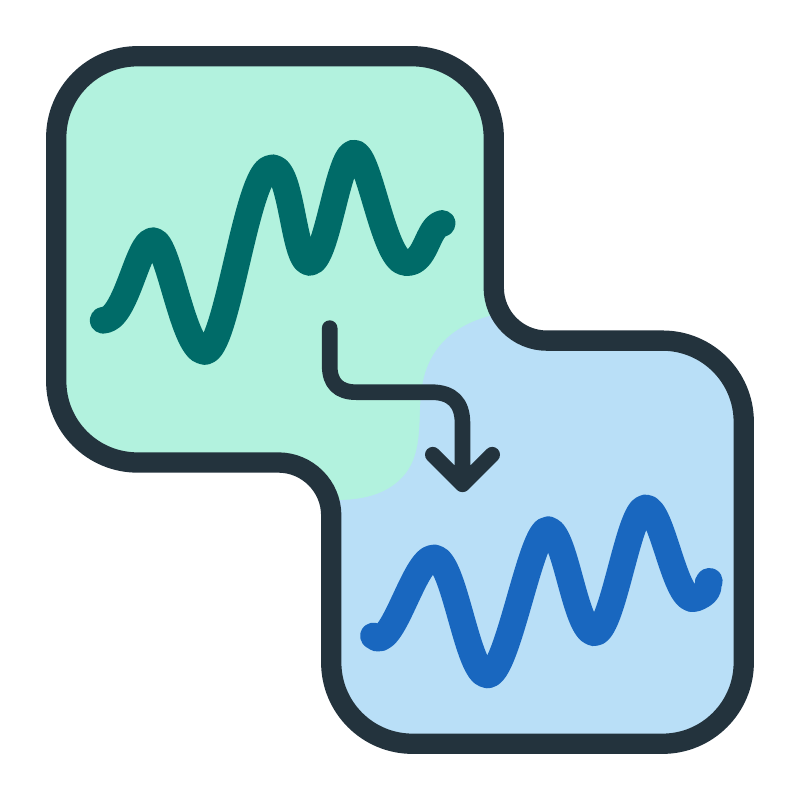}}}
\title{\icon\ XMatch: Enhancing Covariate-Aware Time Series Forecasting through Tree-Structured Exogenous Matching}

\author{\textbf{Ziyang Zhang}\textsuperscript{*}, \textbf{Hanyin Cheng}\textsuperscript{*}, \textbf{Xiangfei Qiu}, \textbf{Yang Shu}\textsuperscript{\Letter}, \textbf{Bin Yang}, \textbf{Chenjuan Guo}\\[0.5em]
East China Normal University\\
\texttt{\{zyzhang11,hycheng,xfqiu\}@stu.ecnu.edu.cn}\\
\texttt{\{yshu,byang,cjguo\}@dase.ecnu.edu.cn}
}

\iclrfinalcopy
\begin{document}

\maketitle
\begingroup
\renewcommand{\thefootnote}{}
\footnotetext[0]{%
$^{*}$ Equal contribution. \quad
\Letter~Corresponding author.
}
\endgroup
\begin{abstract}
Future exogenous variables provide valuable information for forecasting endogenous time series. Existing covariate-aware methods primarily learn the direct influence of exogenous variables on endogenous variables. However, these effects can be complex and change with the pattern of the exogenous variables, making them difficult to capture. Beyond this perspective, we observe that a given exogenous pattern often co-occurs with only a small set of endogenous response patterns. These associations motivate a strategy that matches future and historical exogenous patterns and uses the corresponding endogenous patterns to enhance forecasting. However, in real-world forecasting scenarios with multiple exogenous variables, each exogenous variable provides a distinct dimension for matching, creating a dilemma for this strategy between precise matching and sufficient historical support. To bridge this gap, we propose \textbf{XMatch} (\textbf{EX}ogenous \textbf{MATCH}ing), a covariate-aware forecasting model that realizes the aforementioned strategy through a tree-structured matching process that adaptively adjusts the number of exogenous variables used as matching conditions. Specifically, we first introduce the \textit{ProtoTree Creator}, which organizes historical correspondences between exogenous and endogenous patterns into a \textit{ProtoTree}, whose deeper levels incorporate additional exogenous variables for matching. For forecasting, we then design the \textit{ProtoTree Matcher}, which uses future exogenous variables to query the \textit{ProtoTree} and adaptively determines how many exogenous variables to use for matching based on exogenous pattern similarity and historical support. Finally, the matched endogenous patterns are used as explicit historical evidence to enhance forecasting. Extensive experiments on 12 real-world datasets demonstrate that XMatch outperforms state-of-the-art baselines.
\end{abstract}

\section{Introduction}
Time series forecasting plays a vital role in a wide range of real-world applications, including finance~\citep{sezer2020financial,huang2022dgraph}, transportation~\citep{wu2024autocts++,FOTraj}, healthcare~\citep{wu2025millgnn,miao2021generative}, energy~\citep{wang2026time}, and AIOps~\citep{lin2024cocv,pan2023magicscaler}. Beyond historical target observations, forecasting tasks often involve \emph{covariates} that describe external factors related to the dynamics of the target series~\citep{cheng2026kite, wang2024timexer}. In many real-world scenarios, future covariates are also available or can be estimated in advance using dedicated forecasting systems. For example, weather forecasts provide future temperature and precipitation conditions that are informative for traffic prediction~\citep{GCGNet}. Since these covariates cover the same future horizon as the prediction target, they describe how external factors, such as temperature and precipitation, are expected to evolve during the prediction horizon and can substantially improve forecasting accuracy~\citep{wang2024timexer,zhou2025crosslinear}. Effectively exploiting such future covariate information is therefore important for practical forecasting~\citep{das2023tide,Baguan-TS}.

\begin{figure}[!th]
    \centering
    \includegraphics[width=\linewidth]{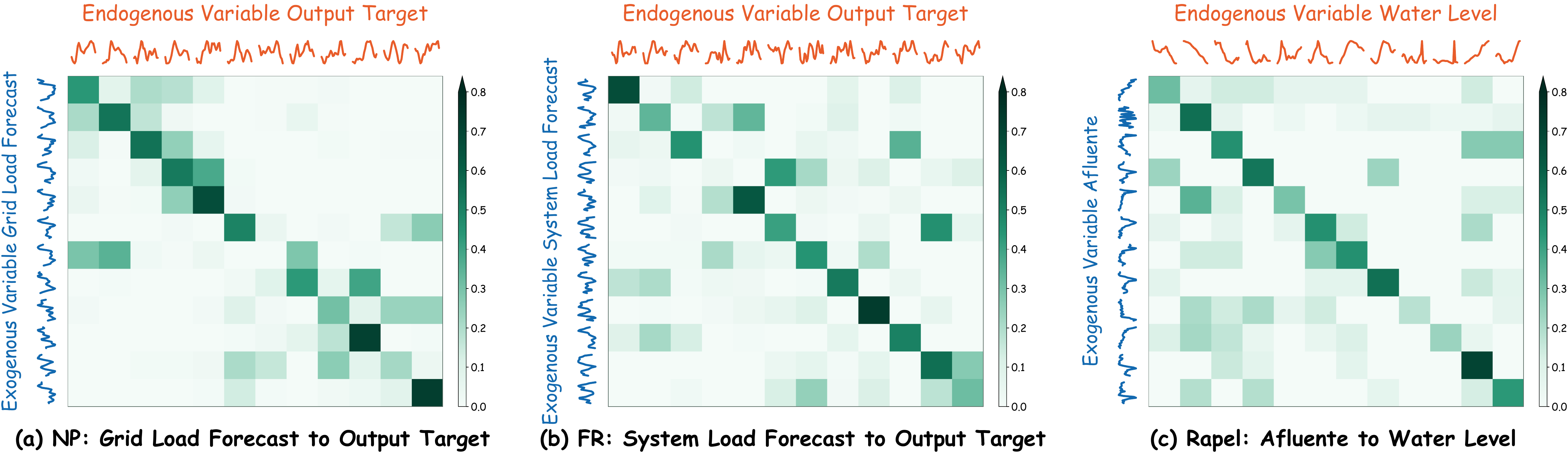}
    \caption{
    Illustration of the frequency of co-occurrences between exogenous and endogenous patterns in real-world datasets. Rows denote exogenous patterns and columns denote endogenous response patterns, while darker cells indicate more co-occurrences. The darker cells in each row are often concentrated in a few columns, showing that a given exogenous pattern tends to co-occur with only a small set of endogenous patterns.  We provide further evidence in Appendix~\ref{app:temporal-recurrence}.}
    \label{fig:intro}
    \vspace{-7mm}
\end{figure}

Existing covariate-aware models primarily learn the direct influence of exogenous variables on endogenous variables~\citep{das2023tide,wang2024timexer,zhou2025crosslinear}.
However, these effects can be complex and change with the pattern of the exogenous variables, making them difficult to capture with a unified interaction mechanism. 
Beyond this conventional perspective, we observe a recurring correspondence between exogenous and endogenous patterns in real-world time series. 
As illustrated in Figure~\ref{fig:intro}, a given exogenous pattern often co-occurs with only a small set of endogenous response patterns.
This suggests that exogenous patterns carry informative signals about which endogenous response patterns are more likely to occur.
Such recurring associations can thus provide a source of information for forecasting through direct pattern matching, reducing reliance on a unified interaction mechanism to capture complex exogenous effects. 
Therefore, in addition to learning exogenous-to-endogenous influence, forecasting models can further leverage historical endogenous responses associated with similar exogenous patterns as explicit evidence for prediction.

Motivated by this observation, we introduce a complementary forecasting perspective that \textbf{explicitly captures recurring endogenous patterns under similar exogenous patterns.} To enhance conventional covariate-aware forecasting, we identify recurring associations between exogenous patterns and their corresponding endogenous responses in historical data, and use these associations to augment prediction. We refer to this strategy as \textit{Exo--Endo Association Augmentation}.

However, this strategy becomes challenging in real-world forecasting scenarios with multiple exogenous variables. \textbf{This strategy faces a fundamental dilemma between precise matching and sufficient historical support.} Each additional exogenous variable requires historical references to match the future exogenous patterns along another dimension. Matching on more variables yields more specific references, but fewer historical samples satisfy these matching requirements. As a result, the resulting endogenous-pattern evidence may be supported by too few samples to be reliable~\citep{DBLP:conf/nips/VandermeulenTA24}. In contrast, matching on fewer variables provides more historical references, but may overlook differences in the remaining variables that are relevant to endogenous prediction~\citep{DBLP:journals/jmlr/GaoH22,DBLP:conf/nips/YangL24}. The key challenge is therefore to identify historical references that closely match the future exogenous patterns while retaining enough samples to support reliable prediction.



To bridge this gap, we propose \textbf{XMatch} (\textbf{EX}ogenous \textbf{MATCH}ing), a covariate-aware forecasting model that realizes the \textit{Exo--Endo Association Augmentation} strategy through a tree-structured matching process that adaptively adjusts the number of exogenous variables used as matching conditions to address the aforementioned dilemma.
Specifically, we first introduce the \textit{ProtoTree Creator} to organize historical Exo--Endo associations into a tree-structured memory, where each deeper level adds one exogenous variable and thus represents a more specific condition with fewer supporting samples. For forecasting, we design the \textit{ProtoTree Matcher}, which uses future exogenous variables to query the \textit{ProtoTree} and adaptively determines the matching depth, assigning more weight to deeper matches only when they are sufficiently similar and well supported; otherwise, the weight remains at shallower, better-supported nodes. Finally, the matched endogenous patterns are aggregated and fused with the forecasting backbone's representations to produce the forecast.

Our contributions are summarized as follows:
\begin{itemize}[labelindent=2mm,leftmargin=5mm]
    \item We reveal recurring correspondences between exogenous and endogenous patterns in covariate-aware forecasting and introduce a new perspective that uses endogenous patterns from historical references with similar exogenous patterns as explicit forecasting evidence.
    \item We propose XMatch, which adaptively adjusts the number of exogenous variables used in tree-structured matching to balance precise matching and sufficient historical support for forecasting.
    \item We conduct extensive experiments on 12 real-world datasets with exogenous variables. The results show that XMatch outperforms state-of-the-art baselines.
\end{itemize}

\section{Related Work}
\subsection{Time Series Forecasting with Exogenous Variables}
Time series forecasting with exogenous variables has long been studied in classical statistics, where ARIMAX~\citep{williams2001multivariate} and SARIMAX~\citep{vagropoulos2016comparison} incorporate external factors through additional regression terms. Recent deep forecasting models provide more flexible mechanisms for covariate integration. NBEATSx~\citep{olivares2023neural} introduces dedicated branches for exogenous inputs, while TiDE~\citep{das2023tide} concatenates static and future covariates with endogenous features. CrossLinear~\citep{zhou2025crosslinear} captures cross-variable dependencies through cross-correlation embeddings and convolutions; TFT~\citep{lim2021temporal}, TimeXer~\citep{wang2024timexer}, and     ExoTST~\citep{tayal2024exotst} employ variable selection or attention mechanisms to integrate exogenous information. More recent approaches further model richer dependencies: DAG~\citep{qiu2026dag} characterizes temporal and channel correlations, GCGNet~\citep{GCGNet} aligns graph-structured correlations, and KITE~\citep{cheng2026kite} introduces knowledge-guided covariate conditioning for probabilistic forecasting. Existing methods primarily model exogenous effects through feature-level interactions, leaving cross-variable pattern correspondences implicit. In contrast, XMatch explicitly retrieves endogenous patterns from historical references with similar exogenous patterns, complementing conventional covariate integration with historical evidence.

\subsection{Retrieval-Augmented Time Series Modeling}
Beyond encoding temporal dynamics solely in model parameters, a growing line of work introduces memory mechanisms to store and reuse features extracted from historical time series. Existing studies in this area mainly follow three directions. \textbf{1) Instance-level retrieval.} RATD~\citep{RATD} retrieves relevant sequences to guide diffusion-based forecasting, while RAFT~\citep{RAFT} augments predictors with the future continuations of similar historical inputs. TS-RAG~\citep{TS-RAG} extends this idea to zero-shot forecasting with time series foundation models, and PFRP~\citep{PFRP} combines predictions retrieved from a global memory bank with those of a local forecaster. \textbf{2) Prototype-based memory.} PUAD~\citep{PUAD} and H-PAD~\citep{H-PAD} compress recurring behavior into prototypical normal patterns for anomaly detection. \textbf{3) Parametric temporal memory.} Other approaches encode historical regularities through learnable memory items~\citep{memto}, multi-resolution attractor memory~\citep{attraos}, or compact temporal bases~\citep{huang2025timebase}. Despite demonstrating the value of explicit pattern reuse, these methods retrieve or summarize patterns from historical observations without explicitly organizing the correspondences between combinations of exogenous patterns and endogenous patterns. In contrast, XMatch organizes these correspondences in a prototype tree and adaptively adjusts the number of exogenous variables used for matching, balancing precise matching with sufficient historical support.

\section{Methodology}
\label{sec:methodology}

In the context of covariate-aware time series forecasting, the historical endogenous time series $X^{\text{endo}} \in \mathbb{R}^{N \times T}$ is accompanied by historical exogenous variables $X^{\text{exo}} \in \mathbb{R}^{D \times T}$ and future exogenous variables $Y^{\text{exo}} \in \mathbb{R}^{D \times F}$, where $N$ is the number of endogenous variables, $D$ is the number of exogenous variables, $T$ is the number of historical time steps, and $F$ is the forecasting horizon. XMatch further constructs a dataset-level \textit{ProtoTree} $\mathcal{T}_{\phi}$ from the training set to encode historical associations between exogenous and endogenous patterns. The objective is to forecast the future endogenous series $\hat{Y}^{\text{endo}} \in \mathbb{R}^{N \times F}$ from the time-series inputs and $\mathcal{T}_{\phi}$:
\begin{align}
\hat{Y}^{\text{endo}} = \mathcal{F}_{\theta}\!\left(X^{\text{endo}}, X^{\text{exo}}, Y^{\text{exo}}; \mathcal{T}_{\phi}\right).
\end{align}
Here, $\mathcal{F}_{\theta}$ is the forecaster parameterized by $\theta$. Each endogenous variable has its own \textit{ProtoTree}; we use one variable as an example below.

\begin{figure*}[t]
    \centering
    \includegraphics[width=\textwidth]{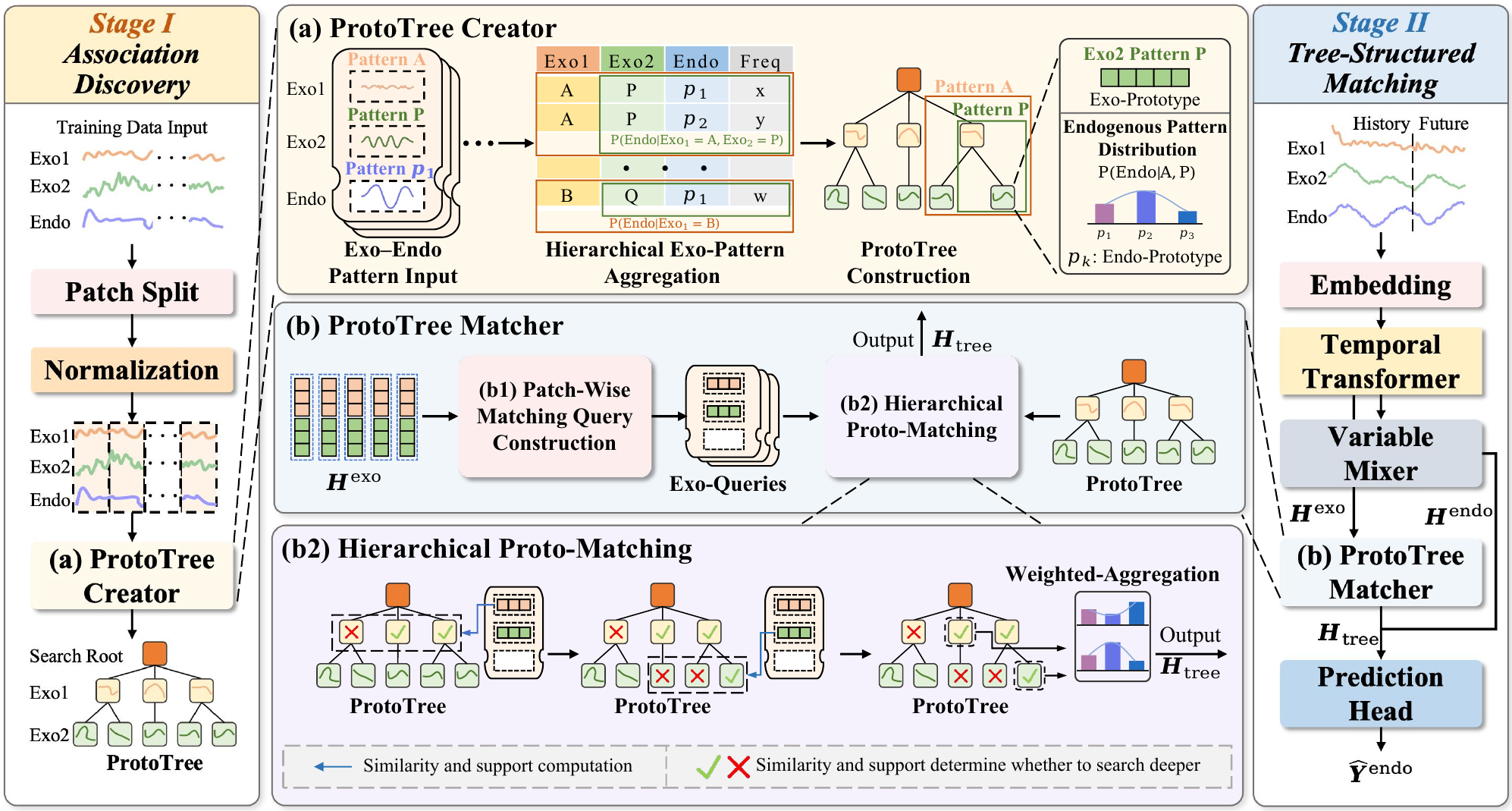}
    \caption{The architecture of XMatch. (a) The \textit{ProtoTree Creator} organizes correspondences between exogenous and endogenous patterns according to the descending discriminative power of exogenous variables. (b) The \textit{ProtoTree Matcher} constructs matching queries from known future exogenous variables and uses \textit{Hierarchical Proto-Matching} to match and aggregate endogenous response prototypes, thereby improving forecasting accuracy.}
    \label{fig:overview}
    \vspace{-5mm}
\end{figure*}

\subsection{Structure Overview}

Figure~\ref{fig:overview} illustrates XMatch, which leverages recurring correspondences between exogenous and endogenous patterns as explicit evidence to enhance forecasting under covariate scenarios, and organizes these associations in a hierarchical \textit{ProtoTree} to address the dilemma between precise matching and sufficient historical support in multi-exogenous matching. Our training procedure consists of two stages. In \textit{\textbf{Stage \uppercase\expandafter{\romannumeral 1\relax}:} Association Discovery}, we mine recurring correspondences between exogenous and endogenous patterns from the training set and organize them into a \textit{ProtoTree} whose deeper nodes match patterns across more exogenous variables, while shallower nodes retain more historical samples to support matching when deeper nodes have too few samples. In \textit{\textbf{Stage \uppercase\expandafter{\romannumeral 2\relax}: }Tree-Structured Matching}, we encode future exogenous variables as matching queries and perform adaptive matching over the \textit{ProtoTree}. The matched response prototypes are then fused with the temporal representation to produce $\hat{Y}^{\mathrm{endo}}$. We detail the two stages below.

\subsection{Stage \uppercase\expandafter{\romannumeral 1\relax}: Association Discovery}
In this stage, we introduce the \textit{ProtoTree Creator} to extract aligned local patterns from the training set, assign discrete pattern labels, and hierarchically aggregate the observed correspondences between exogenous and endogenous patterns into a \textit{ProtoTree}. The resulting tree summarizes reusable association memory for the subsequent matching stage.

\subsubsection{ProtoTree Creator}
As shown in Figure~\ref{fig:overview}a, the \textit{ProtoTree Creator} extracts aligned Exo--Endo patterns, organizes their associations by matching on increasing numbers of exogenous variables, and encodes them as node prototypes for subsequent matching.

\textbf{Aligned Exo--Endo Pattern Input.} To assign discrete pattern labels to the correspondences between exogenous and endogenous patterns observed in the training set, we first extract $M$ aligned patches of length $P$, which may overlap, from each endogenous and exogenous variable in the training set and normalize every patch independently~\cite{patchtst}:
\begin{gather}
\widetilde{E}_j=\operatorname{Norm}(\operatorname{Patchify}(X_j^{\mathrm{exo}})),\quad
\widetilde{S}_i=\operatorname{Norm}(\operatorname{Patchify}(X_i^{\mathrm{endo}})),
\end{gather}
where $\widetilde{E}_j\in\mathbb{R}^{M\times P}$ and $\widetilde{S}_i\in\mathbb{R}^{M\times P}$ denote the normalized patch sequences of the $j$-th exogenous variable and the $i$-th endogenous variable, respectively, with $j\in\{1,\ldots,D\}$ and $i\in\{1,\ldots,N\}$.

To group patterns with similar shapes despite temporal shifts, we cluster the patterns of each variable independently using \textit{Soft-DTW}~\citep{cuturi2017softdtw}, which replaces the hard minimum over temporal alignments with a differentiable soft minimum:
\begin{align}
\operatorname{sDTW}_{\gamma}(\boldsymbol{x},\boldsymbol{c})
=-\gamma\log\sum_{\boldsymbol{A}\in\mathcal{A}}
\exp\!\left(-\frac{\langle\boldsymbol{A},\boldsymbol{\Delta}(\boldsymbol{x},\boldsymbol{c})\rangle}{\gamma}\right),
\end{align}
where a patch is $\boldsymbol{x}\in\mathbb{R}^{P}$, a cluster center is $\boldsymbol{c}\in\mathbb{R}^{P}$, $\mathcal{A}$ is the set of valid alignments and $\boldsymbol{\Delta}$ is the pairwise cost matrix. To accommodate variable-specific pattern diversity, we adopt a \textit{DP-means-style strategy}~\citep{DBLP:conf/icml/KulisJ12} that adaptively determines the number of cluster centers for each variable. After convergence, for each variable $u$, each patch $m$ is assigned a variable-specific pattern label $z_{u,m}\in\{1,\ldots,K_u\}$.

\textbf{Hierarchical Exogenous Pattern Aggregation.} To balance precise matching with sufficient historical support, we construct a hierarchical matching tree, the \textit{ProtoTree}, by first ordering exogenous variables according to their informativeness about endogenous patterns and then progressively combining their pattern labels across increasing numbers of variables.

First, we rank exogenous variables by information gain about endogenous patterns, measured as the reduction in Shannon entropy $H$~\cite{ShannonEntropy}. Let $Z_i^{\mathrm{endo}}$ and $Z_j^{\mathrm{exo}}$ denote the discrete pattern-label variables obtained by clustering endogenous variable $i$ and exogenous variable $j$, respectively. We compute the information gain as follows:
\begin{align}
I_{i,j} = H(Z_i^{\mathrm{endo}}) - H(Z_i^{\mathrm{endo}} \mid Z_j^{\mathrm{exo}}).
\end{align}
For each endogenous variable $i$, we order exogenous variables by descending $I_{i,j}$ from shallow to deep levels. Let $l$ denote tree depth and $\pi_l$ the exogenous variable index at that level.

Second, we combine the ordered labels into a tuple of exogenous pattern labels $\boldsymbol{\eta}_{l,m}=(z^{\mathrm{exo}}_{\pi_1,m},\ldots,z^{\mathrm{exo}}_{\pi_l,m})$ at tree level $l$ for patch $m$. Each distinct label tuple defines a node $v$ that stores all endogenous patterns from matching training patches. Normalizing their pattern-label counts yields $p_v(k)$, the empirical probability of endogenous pattern $k$ for this label tuple:
\begin{align}
p_v(k)=\frac{1}{|\mathcal{M}_v|}
\sum_{m\in\mathcal{M}_v}\mathbb{I}[z_{i,m}^{\mathrm{endo}}=k],
\quad k\in\{1,\ldots,K_i^{\mathrm{endo}}\},
\end{align}
where $\mathcal{M}_v$ contains the training patches whose exogenous pattern labels match the tuple represented by node $v$, and $z_{i,m}^{\mathrm{endo}}$ is the pattern label of endogenous variable $i$ at patch $m$. This hierarchy preserves one-to-many Exo--Endo associations, retaining more historical samples at shallow levels and matching on more exogenous variables at deeper levels.

\textbf{\textit{ProtoTree} Construction.} Each node $v$ at level $l$ represents a tuple of exogenous pattern labels $\boldsymbol{\eta}_{l,m}$ encoded by the tree path leading to it. To support prototype matching, we use an MLP to project the exogenous pattern center $\boldsymbol{c}^{\mathrm{exo}}_{\pi_l,z_v}\in\mathbb{R}^{P}$ associated with its current-level label $z_v=z^{\mathrm{exo}}_{\pi_l,m}$ into a learnable prototype key $\boldsymbol{k}_v\in\mathbb{R}^{d}$. We use another MLP to map the endogenous pattern centers $\boldsymbol{c}^{\mathrm{endo}}_{i,k}\in\mathbb{R}^{P}$, weighted by the node's empirical distribution $p_v(k)$, into a learnable response prototype $\boldsymbol{r}_v\in\mathbb{R}^{d}$ associated with the matched node:
\begin{align}
\boldsymbol{k}_v
&=\operatorname{MLP}_K\!\left(\boldsymbol{c}^{\mathrm{exo}}_{\pi_l,z_v}\right), \qquad
\boldsymbol{r}_v
=\operatorname{MLP}_V\!\left(\sum_k p_v(k)\boldsymbol{c}^{\mathrm{endo}}_{i,k}\right).
\end{align}
This construction grounds prototype initialization in observed patterns and their conditional frequencies, providing an empirical basis for subsequent learning. Finally, we add a virtual \textit{Search Root} and connect it to all first-level nodes, providing a unified entry for hierarchical matching.

\subsection{Stage \uppercase\expandafter{\romannumeral 2\relax}: Tree-Structured Matching}
In this stage, we introduce the \textit{Transformer Backbone} and the \textit{ProtoTree Matcher} to match and aggregate endogenous-response evidence from the \textit{ProtoTree}. The \textit{Transformer Backbone} encodes the inputs into patch-wise representations, using future exogenous latents as matching queries and future endogenous latents as the base forecast. The \textit{ProtoTree Matcher} then performs \textit{Hierarchical Proto-Matching} to adaptively select and aggregate response prototypes, which are fused with the base representation to produce the final forecast.

\subsubsection{Transformer Backbone}
\label{sec:transformer-backbone}

As shown in Figure~\ref{fig:overview}, the Transformer backbone encodes historical observations and known future exogenous variables using the patch-wise representation introduced above~\cite{Triformer}. We append learnable mask tokens $M_i^{\mathrm{endo}}$ to the historical endogenous variables and concatenate the historical exogenous variables with their known future values:
\begin{align}
S^{\mathrm{endo}}_i
=\operatorname{PatchEmbed}([X^{\mathrm{endo}}_i;M^{\mathrm{endo}}_i]),
\qquad
S^{\mathrm{exo}}_j
=\operatorname{PatchEmbed}([X^{\mathrm{exo}}_j;Y^{\mathrm{exo}}_j]).
\end{align}
Let $L$ denote the total number of patches. This produces $S^{\mathrm{endo}}\in\mathbb{R}^{N\times L\times d}$ and $S^{\mathrm{exo}}\in\mathbb{R}^{D\times L\times d}$. When future exogenous variables are unavailable, they are replaced with learnable mask tokens.

We use $\operatorname{Transformer}_{\mathrm{temp}}$ with causal masking to model dependencies among patches, followed by $\operatorname{MLP}_{\mathrm{var}}$ with a residual connection to exchange information across variables:
\begin{gather}
H_{\mathrm{time}}
=\operatorname{Transformer}_{\mathrm{temp}}
\!\left(S;M_{\mathrm{causal}}\right)
\in\mathbb{R}^{(N+D)\times L\times d},\\
H_{\mathrm{mix}}
=H_{\mathrm{time}}
+\operatorname{MLP}_{\mathrm{var}}(H_{\mathrm{time}})
\in\mathbb{R}^{(N+D)\times L\times d},
\end{gather}
where $S=[S^{\mathrm{endo}};S^{\mathrm{exo}}]$ and $M_{\mathrm{causal}}$ blocks attention to subsequent patches. The temporal blocks capture per-variable temporal dependencies, while $\operatorname{MLP}_{\mathrm{var}}$ operates along the variable dimension to exchange information across endogenous and exogenous variables.

We use the future exogenous latents from the temporal Transformer output $H_{\mathrm{time}}$, before variable mixing, as matching queries $H^{\mathrm{query}}$ for the \textit{ProtoTree}.

\subsubsection{\textit{ProtoTree Matcher}}
\label{sec:prototree-matcher}

As shown in Figure~\ref{fig:overview}b, the \textit{ProtoTree Matcher} constructs a matching query for each exogenous variable at each future patch, searches the \textit{ProtoTree} level by level, and aggregates the matched response prototypes as forecasting evidence.

\textbf{Patch-Wise Matching Query Construction.} As shown in Figure~\ref{fig:overview}b1, we construct queries by selecting the future-patch representations of the exogenous variables from the temporal Transformer output $H_{\mathrm{time}}$, before the variable mixer. Let $L_f=\lceil F/P\rceil$ denote the number of future patches. With endogenous variables preceding exogenous variables in the concatenated input, we extract the query for exogenous variable $j$ at future patch $s$ as follows:
\begin{align}
H^{\mathrm{query}}_{j,s,:}
=H_{\mathrm{time}}[N+j,L-L_f+s,:]\in\mathbb{R}^{d}.
\end{align}
This yields $H^{\mathrm{query}}\in\mathbb{R}^{D\times L_f\times d}$. For each future patch, we select these variable-specific queries according to the exogenous variable order $\pi_1,\ldots,\pi_D$ established during tree construction, using the query for variable $\pi_l$ to match the exogenous prototype keys at the corresponding tree level $l$.

\textbf{Hierarchical Proto-Matching.}
\label{sec:hierarchical-proto-matching}
As shown in Figure~\ref{fig:overview}b2, we use the variable-specific future-patch queries constructed by the preceding module to search the \textit{ProtoTree} level by level, with $\boldsymbol{q}_s^{(l)}=H^{\mathrm{query}}_{\pi_l,s,:}$ serving as the query at level $l$.
For each valid exogenous prototype $j$ at this level, we measure its cosine similarity
$a_{s,l,j}=\operatorname{cos}(\boldsymbol{q}_s^{(l)},\boldsymbol{k}_{l,j})$
and convert the similarity scores into relative matching weights using softmax:
\begin{align}
q_{s,l,j}
=\operatorname{softmax}_{j\in\mathcal{J}_l}
\left(a_{s,l,j}/\tau_{\mathrm{route}}\right),
\end{align}
where $\tau_{\mathrm{route}}$ is the routing temperature and $\mathcal{J}_l$ denotes the set of indices of all valid exogenous prototypes at level $l$, not only those of the children observed under the current node $v$. Let $c(v,j)$ denote the corresponding child, if it exists, and let $n_{c(v,j)}$ be the number of training patches matching the full sequence of exogenous pattern labels along its tree path.
During inference, search continues along this branch to the next tree level only when all three conditions below are satisfied:
\begin{align}
b_{s,v,j}
=\mathbb{I}\!\left[
c(v,j)\text{ exists}
\;\land\;
n_{c(v,j)}\geq n_{\min}
\;\land\;
a_{s,l,j}\geq\tau_{\mathrm{sim}}
\right]=1.
\end{align}
Let $\mu_{s,v}$ denote the weight allocated to node $v$ for patch $s$, with the root weight initialized to one for each future patch. The weight allocated to child $c(v,j)$ is then computed as the parent weight multiplied by its matching weight and branch indicator:
\begin{align}
\mu_{s,c(v,j)}
&=\mu_{s,v}\times q_{s,l,j}\times b_{s,v,j}.
\end{align}
Node $v$ retains the weight not passed to its children, yielding the following aggregation weight:
\begin{align}
w_{s,v}
&=\mu_{s,v}\left(1-\sum_{j\in\mathcal{J}_l}q_{s,l,j}b_{s,v,j}\right).
\end{align}
A leaf retains its full allocated weight, and the aggregation weights sum to one. Weights for unsupported, dissimilar, or unobserved branches remain at the parent. This allocation combines evidence from multiple matching branches while favoring matches on more exogenous variables only when they have sufficient historical support and query similarity.

For each future patch $s$, we use the aggregation weights $w_{s,v}$ computed above to combine the node response prototypes $\boldsymbol{r}_v$:
\begin{align}
H_{\mathrm{tree}}[:,s,:]
=\operatorname{Reshape}_{N\times d}\!\left(
\boldsymbol{W}_O\sum_v
w_{s,v}\boldsymbol{r}_v\right),
\end{align}
where the sum runs over all tree nodes except the root, and $\boldsymbol{W}_O\in\mathbb{R}^{Nd\times d}$ projects the aggregated prototype to the endogenous representation space. We stack the outputs across future patches to obtain $H_{\mathrm{tree}}\in\mathbb{R}^{N\times L_f\times d}$, the endogenous-response representation corresponding to the queried future exogenous patterns.

Finally, we add this output to the future endogenous representation $H^{\mathrm{endo}}$ from the \textit{Transformer Backbone} and pass the fused representation through a \textit{Prediction Head} to obtain the final endogenous forecast over the prediction horizon:
\begin{align}
\hat{Y}^{\mathrm{endo}}
=\operatorname{Head}\!\left(H^{\mathrm{endo}}+H_{\mathrm{tree}}\right)
\in\mathbb{R}^{N\times F}.
\end{align}



\section{Experiments}
\label{sec:experiments}
\subsection*{Experimental Setup}
\label{sec:experimental_setup}
\textbf{Datasets.} We evaluate XMatch on 12 real-world datasets under the setting where future exogenous variables are provided as inputs. We conduct both short-term and long-term forecasting experiments on all datasets. For Colbun and Rapel, the short-term setting uses a lookback window of 60 with a prediction horizon of 10, while the long-term setting uses a lookback window of 180 with a prediction horizon of 30. For the other datasets, the corresponding lookback/prediction lengths are $168/24$ and $720/360$. Dataset statistics and variable descriptions are provided in Appendix~\ref{appendix DATASETS}.

\textbf{Baselines.} We compare XMatch with 10 competitive time series baselines. The first group consists of methods that natively model future covariates: DAG~\citep{qiu2026dag}, KITE~\citep{cheng2026kite}, GCGNet~\citep{GCGNet}, TimeXer~\citep{wang2024timexer}, TFT~\citep{lim2021temporal}, and TiDE~\citep{das2023tide}. The second group contains strong general-purpose time series forecasters: DUET~\citep{qiu2025duet}, CrossLinear~\citep{zhou2025crosslinear}, Amplifier~\citep{amplifier}, and TimeKAN~\citep{Timekan}. For a fair comparison, models in the second group are equipped with the same MLP-based future-covariate fusion strategy used in prior covariate forecasting studies~\citep{qiu2026dag} as described in Appendix~\ref{appendix MLP}.

\textbf{Implementation Details.} To ensure consistency with previous studies, we adopt Mean Squared Error (MSE) and Mean Absolute Error (MAE) as evaluation metrics, where lower values indicate better performance. We use the TFB benchmark~\citep{qiu2024tfb} for unified evaluation, and all baseline results are obtained within this framework. Evaluation retains the final incomplete batch to ensure that all test windows are included. Further implementation details are provided in Appendix~\ref{app:implementation-details}.

\subsection{Main Results}
\label{sec:main_results}

\begin{table*}[!t]
\centering
\renewcommand{\arraystretch}{1.2}
\caption{Average results on 12 real-world datasets, where the inputs are $X^{\text{endo}}$, $X^{\text{exo}}$, and $Y^{\text{exo}}$. \textcolor{red}{\textbf{Red}}: the best, \textcolor{blue}{\underline{Blue}}: the 2nd best. Full results with $Y^{\text{exo}}$ are reported in Table~\ref{tab:main_results_full}.}
\label{tab:main_results_avg}
\setlength{\tabcolsep}{4.2pt}
\resizebox{\textwidth}{!}{%
\begin{tabular}{c|cc|cc|cc|cc|cc|cc|cc|cc|cc|cc|cc}
\toprule
Models & \multicolumn{2}{c}{XMatch} & \multicolumn{2}{c}{DAG} & \multicolumn{2}{c}{KITE} & \multicolumn{2}{c}{GCGNet} & \multicolumn{2}{c}{TimeXer} & \multicolumn{2}{c}{TFT} & \multicolumn{2}{c}{TiDE} & \multicolumn{2}{c}{DUET} & \multicolumn{2}{c}{CrossLinear} & \multicolumn{2}{c}{Amplifier} & \multicolumn{2}{c}{TimeKAN} \\
Metrics & \multicolumn{1}{c}{mse} & \multicolumn{1}{c}{mae} & \multicolumn{1}{c}{mse} & \multicolumn{1}{c}{mae} & \multicolumn{1}{c}{mse} & \multicolumn{1}{c}{mae} & \multicolumn{1}{c}{mse} & \multicolumn{1}{c}{mae} & \multicolumn{1}{c}{mse} & \multicolumn{1}{c}{mae} & \multicolumn{1}{c}{mse} & \multicolumn{1}{c}{mae} & \multicolumn{1}{c}{mse} & \multicolumn{1}{c}{mae} & \multicolumn{1}{c}{mse} & \multicolumn{1}{c}{mae} & \multicolumn{1}{c}{mse} & \multicolumn{1}{c}{mae} & \multicolumn{1}{c}{mse} & \multicolumn{1}{c}{mae} & \multicolumn{1}{c}{mse} & \multicolumn{1}{c}{mae} \\
\midrule
NP & \textcolor{red}{\textbf{0.282}} & \textcolor{red}{\textbf{0.299}} & 0.362 & 0.344 & \textcolor{blue}{\underline{0.325}} & \textcolor{blue}{\underline{0.323}} & 0.370 & 0.348 & 0.418 & 0.371 & 0.379 & 0.375 & 0.443 & 0.400 & 0.411 & 0.408 & 0.371 & 0.387 & 0.420 & 0.418 & 0.405 & 0.419 \\
\midrule
PJM & \textcolor{red}{\textbf{0.085}} & \textcolor{red}{\textbf{0.178}} & \textcolor{blue}{\underline{0.093}} & 0.180 & 0.096 & \textcolor{blue}{\underline{0.179}} & 0.095 & 0.187 & 0.108 & 0.198 & 0.114 & 0.207 & 0.142 & 0.246 & 0.102 & 0.197 & 0.112 & 0.223 & 0.137 & 0.246 & 0.139 & 0.262 \\
\midrule
BE & \textcolor{red}{\textbf{0.420}} & \textcolor{red}{\textbf{0.273}} & \textcolor{blue}{\underline{0.423}} & \textcolor{blue}{\underline{0.280}} & 0.428 & 0.286 & 0.431 & 0.294 & 0.452 & 0.290 & 0.454 & 0.291 & 0.498 & 0.325 & 0.515 & 0.354 & 0.479 & 0.337 & 0.559 & 0.413 & 0.548 & 0.407 \\
\midrule
FR & \textcolor{blue}{\underline{0.410}} & \textcolor{red}{\textbf{0.216}} & 0.414 & \textcolor{blue}{\underline{0.219}} & \textcolor{red}{\textbf{0.387}} & 0.225 & 0.415 & 0.234 & 0.427 & 0.241 & 0.504 & 0.257 & 0.484 & 0.281 & 0.496 & 0.327 & 0.483 & 0.298 & 0.554 & 0.408 & 0.547 & 0.374 \\
\midrule
DE & \textcolor{red}{\textbf{0.349}} & \textcolor{red}{\textbf{0.369}} & 0.370 & \textcolor{blue}{\underline{0.370}} & \textcolor{blue}{\underline{0.350}} & \textcolor{blue}{\underline{0.370}} & 0.401 & 0.389 & 0.475 & 0.418 & 0.489 & 0.446 & 0.499 & 0.447 & 0.482 & 0.430 & 0.485 & 0.452 & 0.473 & 0.441 & 0.473 & 0.445 \\
\midrule
Energy & \textcolor{red}{\textbf{0.089}} & \textcolor{red}{\textbf{0.231}} & 0.124 & 0.267 & \textcolor{blue}{\underline{0.111}} & \textcolor{blue}{\underline{0.257}} & 0.131 & 0.277 & 0.163 & 0.315 & 0.130 & 0.283 & 0.153 & 0.302 & 0.203 & 0.367 & 0.239 & 0.402 & 0.233 & 0.389 & 0.218 & 0.381 \\
\midrule
Sdwpfm1 & \textcolor{red}{\textbf{0.402}} & \textcolor{red}{\textbf{0.436}} & 0.423 & 0.461 & \textcolor{blue}{\underline{0.417}} & \textcolor{blue}{\underline{0.451}} & 0.424 & 0.457 & 0.701 & 0.609 & 0.482 & 0.474 & 0.483 & 0.507 & 0.599 & 0.570 & 0.426 & 0.502 & 0.437 & 0.490 & 0.447 & 0.534 \\
\midrule
Sdwpfm2 & \textcolor{red}{\textbf{0.469}} & \textcolor{red}{\textbf{0.479}} & 0.477 & 0.485 & \textcolor{blue}{\underline{0.475}} & \textcolor{blue}{\underline{0.483}} & \textcolor{blue}{\underline{0.475}} & 0.486 & 0.803 & 0.653 & 0.476 & 0.488 & 0.486 & 0.516 & 0.514 & 0.490 & 0.533 & 0.573 & 0.491 & 0.512 & 0.497 & 0.564 \\
\midrule
Sdwpfh1 & \textcolor{red}{\textbf{0.406}} & \textcolor{red}{\textbf{0.442}} & 0.448 & \textcolor{blue}{\underline{0.486}} & \textcolor{blue}{\underline{0.441}} & 0.491 & 0.450 & 0.500 & 0.746 & 0.643 & 0.479 & 0.491 & 0.453 & 0.508 & 0.539 & 0.516 & 0.557 & 0.593 & 0.537 & 0.598 & 0.577 & 0.638 \\
\midrule
Sdwpfh2 & \textcolor{red}{\textbf{0.413}} & \textcolor{red}{\textbf{0.451}} & 0.523 & 0.530 & \textcolor{blue}{\underline{0.501}} & 0.527 & 0.520 & 0.536 & 0.891 & 0.719 & 0.566 & \textcolor{blue}{\underline{0.521}} & 0.599 & 0.583 & 0.647 & 0.566 & 0.538 & 0.574 & 0.521 & 0.581 & 0.647 & 0.672 \\
\midrule
Colbun & \textcolor{blue}{\underline{0.098}} & \textcolor{red}{\textbf{0.149}} & \textcolor{blue}{\underline{0.098}} & \textcolor{blue}{\underline{0.154}} & \textcolor{red}{\textbf{0.088}} & 0.172 & 0.107 & 0.175 & 0.145 & 0.235 & 0.238 & 0.297 & 0.164 & 0.227 & 0.198 & 0.266 & 0.126 & 0.195 & 0.173 & 0.246 & 0.128 & 0.175 \\
\midrule
Rapel & \textcolor{blue}{\underline{0.238}} & \textcolor{red}{\textbf{0.284}} & \textcolor{red}{\textbf{0.230}} & 0.305 & 0.244 & \textcolor{blue}{\underline{0.295}} & 0.306 & 0.307 & 0.344 & 0.362 & 0.305 & 0.333 & 0.320 & 0.351 & 0.269 & 0.326 & 0.252 & 0.313 & 0.257 & 0.321 & 0.249 & 0.311 \\
\midrule
1\textsuperscript{st} Count & \textcolor{red}{\textbf{9}} & \textcolor{red}{\textbf{12}} & 1 & 0 & \textcolor{blue}{\underline{2}} & 0 & 0 & 0 & 0 & 0 & 0 & 0 & 0 & 0 & 0 & 0 & 0 & 0 & 0 & 0 & 0 & 0 \\
\bottomrule
\end{tabular}%
}
\end{table*}

Table~\ref{tab:main_results_avg} presents the average forecasting results on 12 real-world datasets when future exogenous variables are available. Full results for both forecasting horizons are reported in Appendix~\ref{app:forecasting_results}. We have the following observations:
1) XMatch achieves the highest number of first-place rankings, with 9 in MSE and 12 in MAE. It obtains the lowest values for both metrics on nine datasets, outperforming both forecasters designed for future covariates and general-purpose forecasters augmented with future-covariate fusion.
2) The improvements are particularly pronounced on Energy and Sdwpfh2. Compared with the strongest baseline for each metric, XMatch reduces MSE and MAE by 20\% and 10\% on Energy, and by 18\% and 13\% on Sdwpfh2, respectively. These results support the effectiveness of retrieving endogenous patterns from historical references with similar exogenous patterns, as further illustrated in Appendix~\ref{app:forecasting_results}.

\subsection{Model Analysis}
\label{sec:model_analysis}

\begin{wraptable}{r}{0.60\textwidth}
\renewcommand{\arraystretch}{1.0}
\centering
\caption{Average results of ablation studies for XMatch.}

\label{tab:ablation_studies}
\fontsize{6.5}{7.5}\selectfont
\setlength{\tabcolsep}{2pt}
\begin{tabularx}{\linewidth}{@{}c|*{2}{>{\centering\arraybackslash}X}|*{2}{>{\centering\arraybackslash}X}|*{2}{>{\centering\arraybackslash}X}@{}}
\toprule
Dataset & \multicolumn{2}{c|}{BE} & \multicolumn{2}{c|}{DE} & \multicolumn{2}{c}{Sdwpfh1} \\ \midrule
Metrics & mse & mae & mse & mae & mse & mae \\ \midrule
(a) w/o tree retrieval & 0.440 & 0.298 & 0.371 & 0.373 & 0.442 & 0.477 \\ \midrule
(b) w/o variable ordering & 0.438 & 0.289 & 0.410 & 0.396 & 0.457 & 0.481 \\ \midrule
(c) w/o search stopping & 0.430 & 0.283 & 0.380 & 0.375 & 0.419 & 0.454 \\ \midrule
(d) endo history retrieval & 0.433 & 0.299 & 0.356 & 0.373 & 0.478 & 0.491 \\ \midrule
(e) mix exo retrieval & 0.425 & 0.285 & 0.363 & 0.377 & 0.470 & 0.487 \\ \midrule
\textbf{XMatch (Full)} & \textbf{0.420} & \textbf{0.273} & \textbf{0.349} & \textbf{0.369} & \textbf{0.406} & \textbf{0.442} \\
\bottomrule
\end{tabularx}
\end{wraptable}

\textbf{Ablation Studies.}
\label{sec:ablation_studies}
To assess the contribution of the components of XMatch, we compare the full model with five variants in Table~\ref{tab:ablation_studies}:
(a) \emph{w/o tree retrieval} disables the retrieval branch and sets its residual contribution to zero. Its degraded performance indicates that retrieved endogenous response patterns provide useful evidence.
(b) \emph{w/o variable ordering} retains the full model's clusters and exogenous variable set but rebuilds the prefix tree using a fixed random permutation instead of information-gain ordering. The performance drop supports prioritizing informative exogenous variables to guide hierarchical matching.
(c) \emph{w/o search stopping} disables similarity-based and reliability-based early stopping during training and inference, continuing along existing branches until a leaf or a missing branch is reached. The resulting degradation highlights the role of adaptive stopping in avoiding dissimilar or insufficiently supported matches.
(d) \emph{endo history retrieval} replaces tree retrieval with the average subsequent endogenous sequence of the five nearest training records, queried using the historical endogenous context. Its inferior results support using future exogenous patterns to retrieve relevant endogenous responses.
(e) \emph{mix exo retrieval} queries the training library using concatenated, channel-wise normalized future exogenous patches and averages the endogenous patches of the five nearest records, without clustering or a tree. Its lower accuracy supports organizing exogenous--endogenous associations into \textit{ProtoTree} for effective matching. 
Ultimately, the full XMatch model, with variable ordering and adaptive matching, balances matching precision and historical support and achieves strong results.

\begin{figure}[!t]
    \centering
    \begin{subfigure}[t]{0.24\textwidth}
        \centering
        \includegraphics[width=\linewidth]{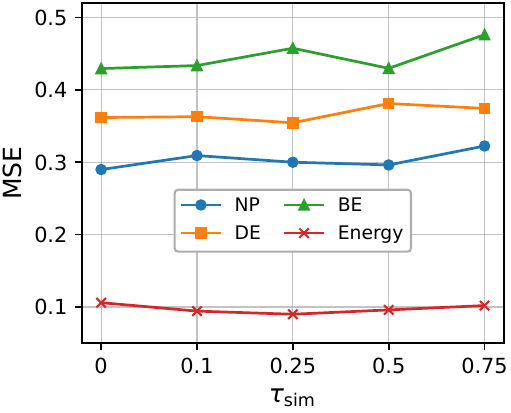}
        \caption{Similarity threshold}
        \label{fig:sensitivity-similarity}
    \end{subfigure}\hfill
    \begin{subfigure}[t]{0.24\textwidth}
        \centering
        \includegraphics[width=\linewidth]{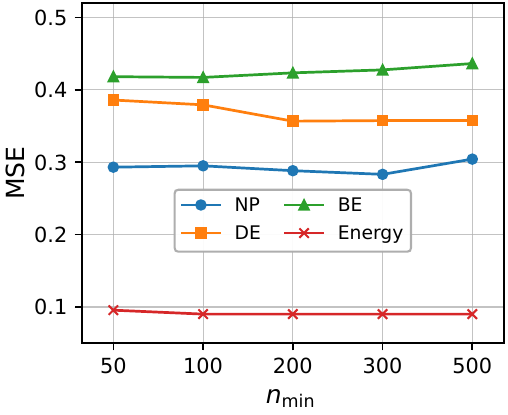}
        \caption{Minimum support}
        \label{fig:sensitivity-support}
    \end{subfigure}\hfill
    \begin{subfigure}[t]{0.24\textwidth}
        \centering
        \includegraphics[width=\linewidth]{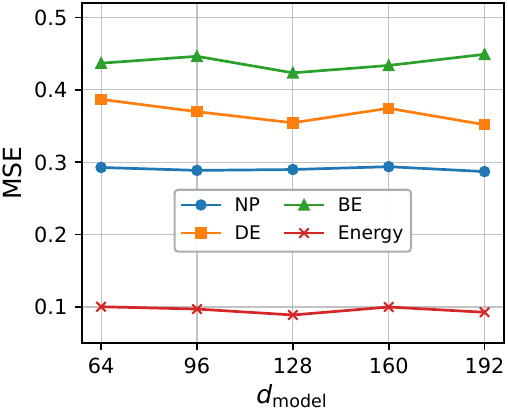}
        \caption{Model dimension}
        \label{fig:sensitivity-dimension}
    \end{subfigure}\hfill
    \begin{subfigure}[t]{0.24\textwidth}
        \centering
        \includegraphics[width=\linewidth]{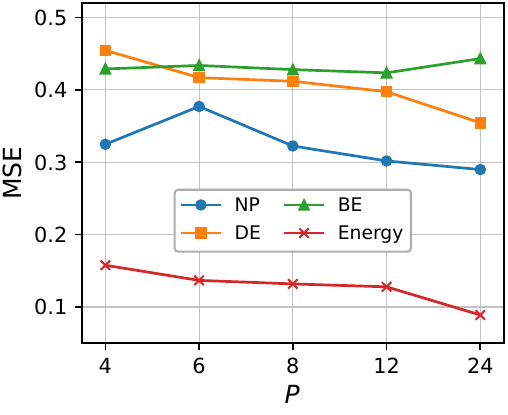}
        \caption{Patch length}
        \label{fig:sensitivity-patch}
    \end{subfigure}
    \caption{Hyperparameter sensitivity of XMatch on NP, DE, BE, and Energy.}
    \label{fig:hyperparameter-sensitivity}
\end{figure}

\textbf{Parameter Sensitivity.}
Sensitivity studies of XMatch yield the following observations:
1) Figures~\ref{fig:sensitivity-similarity} and~\ref{fig:sensitivity-support} show how performance varies with the similarity threshold $\tau_{\mathrm{sim}}$ and minimum support count $n_{\min}$. A stricter similarity threshold does not necessarily improve forecasting accuracy. An excessively large $n_{\min}$ may reduce the number of retrieved matches, whereas an excessively small value may admit matches with insufficient historical support, reducing the reliability of the retrieved evidence.
2) Figure~\ref{fig:sensitivity-dimension} shows that XMatch is relatively insensitive to the model dimension $d$, while smaller dimensions help reduce computational cost. We recommend selecting $d$ within $[96, 160]$.
3) Figure~\ref{fig:sensitivity-patch} explores the effect of patch length on forecasting performance. The optimal patch length varies across datasets due to differences in temporal dependencies; however, a patch size in $[12, 24]$ tends to yield favorable results. Very small patch lengths may increase computational complexity, while excessively large patch lengths may weaken the model's ability to capture local dependencies. Additional sensitivity analyses are provided in Appendix~\ref{app:parameter-sensitivity}.

\textbf{Matching Precision and Historical Support Analysis.}
We compare \textit{ProtoTree} with direct retrieval using increasing numbers of exogenous variables on Energy and Sdwpfh1. Weighted DTW-based similarity between retrieved and actual future endogenous patterns measures match quality, while effective historical support assesses evidence sufficiency. Details are provided in Appendix~\ref{app:matching-analysis}.

\begin{figure}[!t]
    \centering
    \begin{subfigure}[t]{0.24\textwidth}
        \centering
        \includegraphics[width=\linewidth]{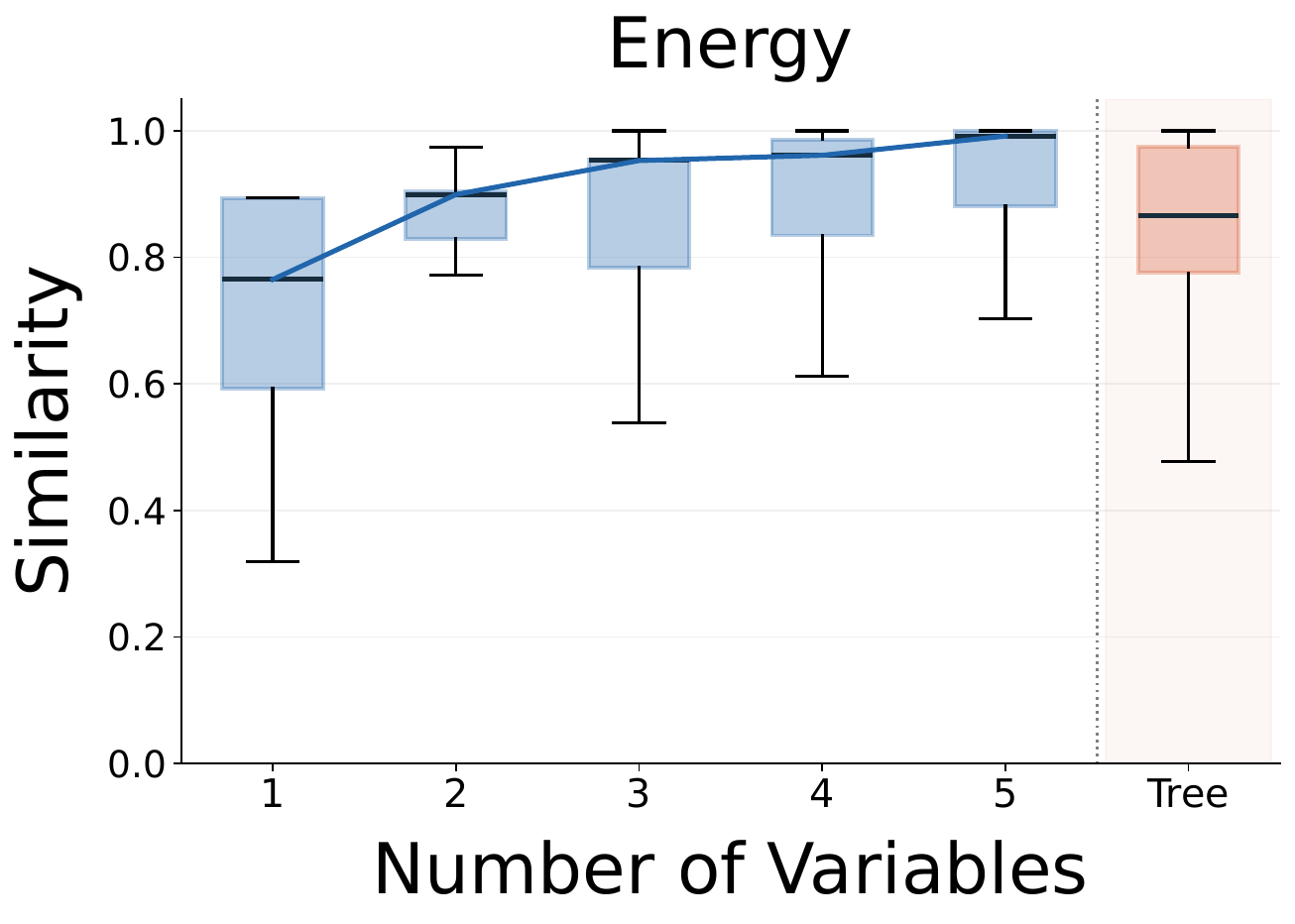}
        \caption{Energy: Similarity}
        \label{fig:retrieval-energy-similarity}
    \end{subfigure}\hfill
    \begin{subfigure}[t]{0.24\textwidth}
        \centering
        \includegraphics[width=\linewidth]{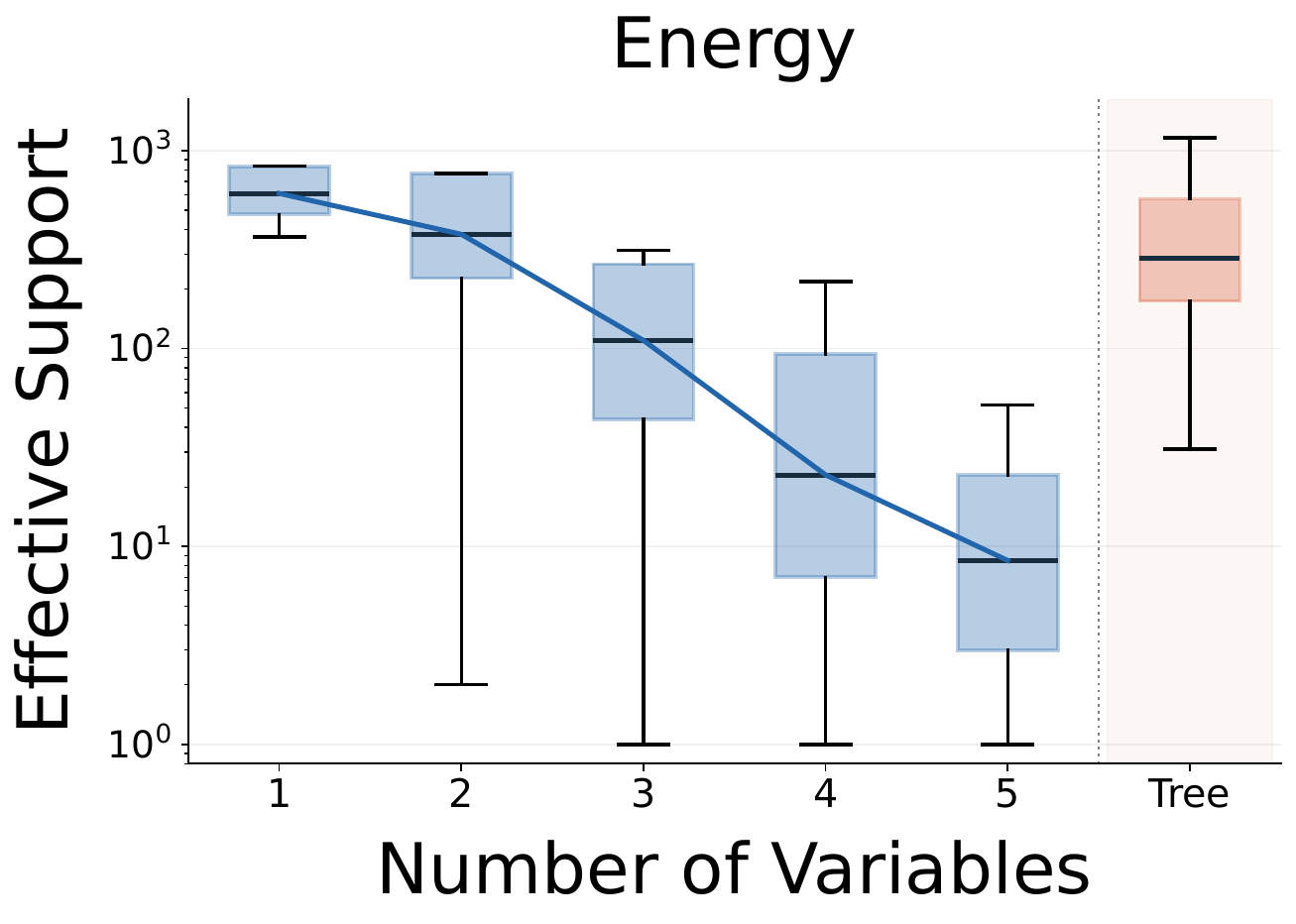}
        \caption{Energy: Support}
        \label{fig:retrieval-energy-support}
    \end{subfigure}\hfill
    \begin{subfigure}[t]{0.24\textwidth}
        \centering
        \includegraphics[width=\linewidth]{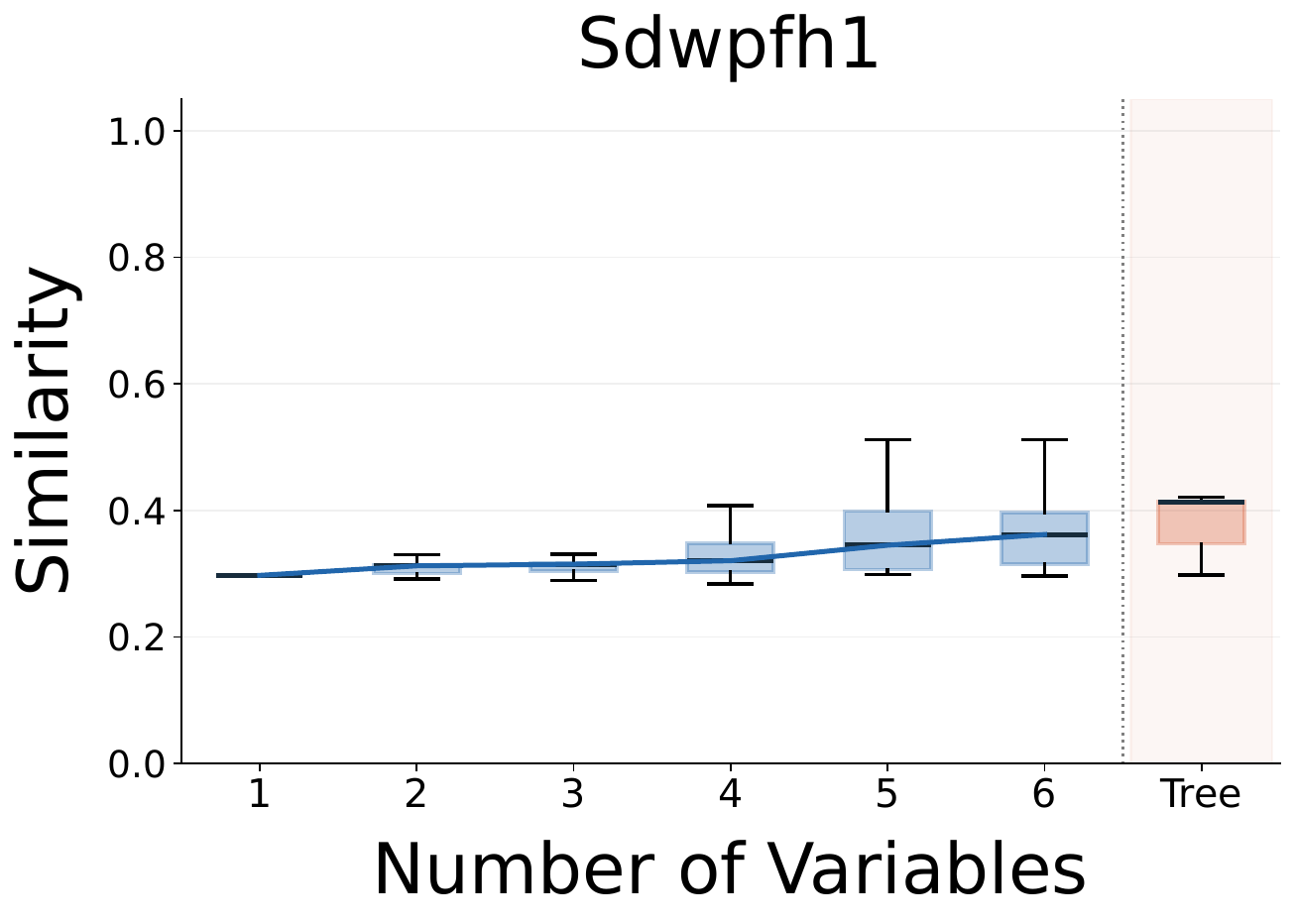}
        \caption{Sdwpfh1: Similarity}
        \label{fig:retrieval-sdwpfh1-similarity}
    \end{subfigure}\hfill
    \begin{subfigure}[t]{0.24\textwidth}
        \centering
        \includegraphics[width=\linewidth]{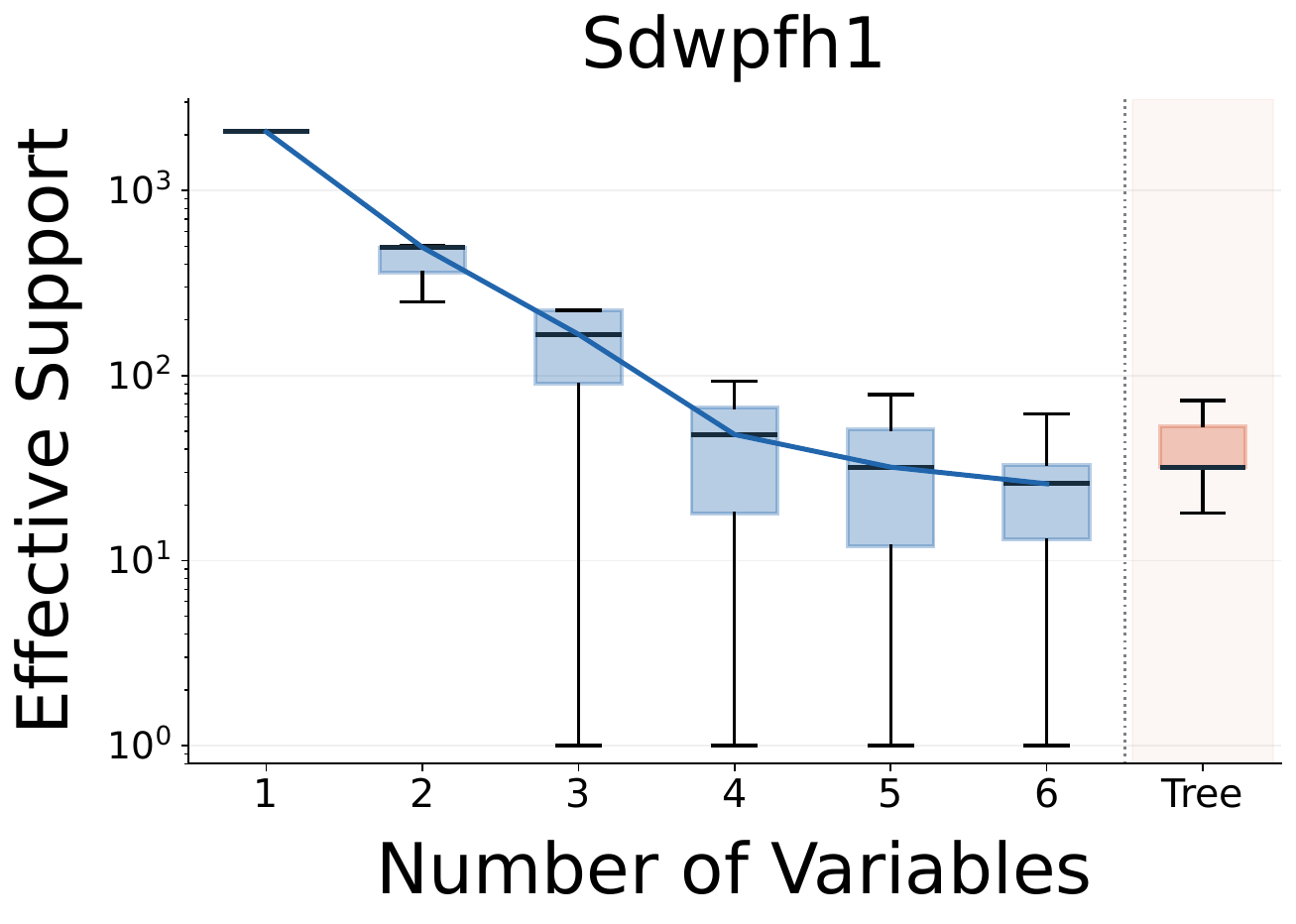}
        \caption{Sdwpfh1: Support}
        \label{fig:retrieval-sdwpfh1-support}
    \end{subfigure}
    \caption{Retrieval similarity and effective historical support on Energy and Sdwpfh1. Support is shown on a logarithmic scale.}
    \label{fig:retrieval-specificity-reliability}
\end{figure}

As shown in Figure~\ref{fig:retrieval-specificity-reliability}, using more exogenous variables in direct retrieval improves similarity but reduces historical support. \textit{ProtoTree} achieves high similarity while retaining greater support than direct retrieval using all exogenous variables, balancing matching precision and historical support. This balance reduces reliance on too few historical examples for highly specific exogenous patterns.

\section{Conclusion}
In this work, we propose XMatch, a covariate-aware forecasting framework that exploits recurring correspondences between exogenous and endogenous patterns as explicit forecasting evidence. The \textit{ProtoTree Creator} organizes historical correspondences between exogenous and endogenous patterns into a hierarchical prototype tree. The \textit{ProtoTree Matcher} then matches future exogenous representations against exogenous prototype keys and aggregates the corresponding endogenous response prototypes, balancing precise matching with sufficient historical support. Experiments on 12 real-world datasets demonstrate the strong overall performance of XMatch against state-of-the-art forecasting baselines.

\clearpage

\bibliography{reference}

\begin{thebibliography}{96}
\providecommand{\natexlab}[1]{#1}
\providecommand{\url}[1]{\texttt{#1}}
\expandafter\ifx\csname urlstyle\endcsname\relax
  \providecommand{\doi}[1]{doi: #1}\else
  \providecommand{\doi}{doi: \begingroup \urlstyle{rm}\Url}\fi

\bibitem[Box \& Pierce(1970)Box and Pierce]{box1970distribution}
George~EP Box and David~A Pierce.
\newblock Distribution of residual autocorrelations in
  autoregressive-integrated moving average time series models.
\newblock \emph{Journal of the American statistical Association}, 65\penalty0
  (332):\penalty0 1509--1526, 1970.

\bibitem[Chen et~al.(2023)Chen, Chen, Shang, Wu, Zheng, Wen, and Zhang]{MAGNN}
Ling Chen, Donghui Chen, Zongjiang Shang, Binqing Wu, Cen Zheng, Bo~Wen, and
  Wei Zhang.
\newblock Multi-scale adaptive graph neural network for multivariate time
  series forecasting.
\newblock \emph{TKDE}, 35\penalty0 (10):\penalty0 10748--10761, 2023.

\bibitem[Cheng et~al.(2026{\natexlab{a}})Cheng, Liu, and
  Lv]{cheng2026metagnsdformer}
Fang Cheng, Hui Liu, and Xinwei Lv.
\newblock Metagnsdformer: Meta-learning enhanced gated non-stationary informer
  with frequency-aware attention for point-interval remaining useful life
  prediction of lithium-ion batteries.
\newblock \emph{Advanced Engineering Informatics}, 69:\penalty0 103798,
  2026{\natexlab{a}}.

\bibitem[Cheng et~al.(2026{\natexlab{b}})Cheng, Wu, Qiu, Shu, Yang, and
  Guo]{cheng2026ccd}
Hanyin Cheng, Xingjian Wu, Xiangfei Qiu, Yang Shu, Bin Yang, and Chenjuan Guo.
\newblock {CCD}: Capturing cross-correlations with deformable convolutional
  networks for multivariate time series forecasting.
\newblock In \emph{KDD}, 2026{\natexlab{b}}.

\bibitem[Cheng et~al.(2026{\natexlab{c}})Cheng, Wu, Shu, Rao, Pan, Yang, and
  Guo]{cheng2026cora}
Hanyin Cheng, Xingjian Wu, Yang Shu, Zhongwen Rao, Lujia Pan, Bin Yang, and
  Chenjuan Guo.
\newblock Cora: Boosting time series foundation models for multivariate
  forecasting through correlation-aware adapter.
\newblock In \emph{ICLR}, 2026{\natexlab{c}}.

\bibitem[Cheng et~al.(2026{\natexlab{d}})Cheng, Zhang, Lu, Chen, Wang, Shu,
  Yang, and Guo]{STAR}
Hanyin Cheng, Ruitong Zhang, Yuning Lu, Peng Chen, Meng Wang, Yang Shu, Bin
  Yang, and Chenjuan Guo.
\newblock {STAR}: Boosting time series foundation models for anomaly detection
  through state-aware adapter.
\newblock In \emph{NeurIPS}, 2026{\natexlab{d}}.

\bibitem[Cheng et~al.(2026{\natexlab{e}})Cheng, Zhou, Shu, and
  Guo]{cheng2026kite}
Hanyin Cheng, Jingrong Zhou, Yang Shu, and Chenjuan Guo.
\newblock Kite: Knowledge-guided probabilistic modeling for time series
  forecasting with exogenous variables.
\newblock In \emph{ICML}, 2026{\natexlab{e}}.

\bibitem[Cirstea et~al.(2022)Cirstea, Guo, Yang, Kieu, Dong, and
  Pan]{Triformer}
Razvan{-}Gabriel Cirstea, Chenjuan Guo, Bin Yang, Tung Kieu, Xuanyi Dong, and
  Shirui Pan.
\newblock Triformer: Triangular, variable-specific attentions for long sequence
  multivariate time series forecasting.
\newblock In \emph{IJCAI}, pp.\  1994--2001, 2022.

\bibitem[Cuturi \& Blondel(2017)Cuturi and Blondel]{cuturi2017softdtw}
Marco Cuturi and Mathieu Blondel.
\newblock Soft-dtw: a differentiable loss function for time-series.
\newblock In \emph{ICML}, volume~70, pp.\  894--903, 2017.

\bibitem[Das et~al.(2023)Das, Kong, Leach, Mathur, Sen, and Yu]{das2023tide}
Abhimanyu Das, Weihao Kong, Andrew Leach, Shaan Mathur, Rajat Sen, and Rose Yu.
\newblock Long-term forecasting with tide: Time-series dense encoder.
\newblock \emph{arXiv preprint arXiv:2304.08424}, 2023.

\bibitem[Du et~al.(2026)Du, Han, and Guo]{PFRP}
Dazhao Du, Tao Han, and Song Guo.
\newblock Predicting the future by retrieving the past.
\newblock In \emph{AAAI}, pp.\  20896--20904, 2026.

\bibitem[Ekambaram et~al.(2024)Ekambaram, Jati, Dayama, Mukherjee, Nguyen,
  Gifford, Reddy, and Kalagnanam]{TTM}
Vijay Ekambaram, Arindam Jati, Pankaj Dayama, Sumanta Mukherjee, Nam Nguyen,
  Wesley~M. Gifford, Chandra Reddy, and Jayant Kalagnanam.
\newblock Tiny time mixers (ttms): Fast pre-trained models for enhanced
  zero/few-shot forecasting of multivariate time series.
\newblock In \emph{NeurIPS}, 2024.

\bibitem[Fang et~al.(2024)Fang, Xie, Zhao, Chen, Gao, and
  Zheng]{fang2024temporal}
Yuchen Fang, Jiandong Xie, Yan Zhao, Lu~Chen, Yunjun Gao, and Kai Zheng.
\newblock Temporal-frequency masked autoencoders for time series anomaly
  detection.
\newblock In \emph{ICDE}, pp.\  1228--1241, 2024.

\bibitem[Fei et~al.(2025)Fei, Yi, Fan, Zhang, and Niu]{amplifier}
Jingru Fei, Kun Yi, Wei Fan, Qi~Zhang, and Zhendong Niu.
\newblock Amplifier: Bringing attention to neglected low-energy components in
  time series forecasting.
\newblock In \emph{AAAI}, volume~39, pp.\  11645--11653, 2025.

\bibitem[Gao \& Hastie(2022)Gao and Hastie]{DBLP:journals/jmlr/GaoH22}
Zijun Gao and Trevor Hastie.
\newblock Lincde: Conditional density estimation via lindsey's method.
\newblock \emph{JMLR}, 23:\penalty0 52:1--52:55, 2022.

\bibitem[Han et~al.(2025)Han, Lee, Cha, Arik, and Yoon]{RAFT}
Sungwon Han, Seungeon Lee, Meeyoung Cha, Sercan~{"{O}}. Arik, and Jinsung Yoon.
\newblock Retrieval augmented time series forecasting.
\newblock In \emph{ICML}, volume 267, 2025.

\bibitem[Hersbach et~al.(2020)Hersbach, Bell, Berrisford, Hirahara,
  Hor{\'a}nyi, Mu{\~n}oz-Sabater, Nicolas, Peubey, Radu, Schepers,
  et~al.]{era5}
Hans Hersbach, Bill Bell, Paul Berrisford, Shoji Hirahara, Andr{\'a}s
  Hor{\'a}nyi, Joaqu{\'\i}n Mu{\~n}oz-Sabater, Julien Nicolas, Carole Peubey,
  Raluca Radu, Dinand Schepers, et~al.
\newblock The era5 global reanalysis.
\newblock \emph{Quarterly journal of the royal meteorological society},
  146\penalty0 (730):\penalty0 1999--2049, 2020.

\bibitem[Hu et~al.(2024{\natexlab{a}})Hu, Hu, Chen, Jin, Pan, Wen, and
  Liang]{attraos}
Jiaxi Hu, Yuehong Hu, Wei Chen, Ming Jin, Shirui Pan, Qingsong Wen, and Yuxuan
  Liang.
\newblock Attractor memory for long-term time series forecasting: {A} chaos
  perspective.
\newblock In \emph{NeurIPS}, 2024{\natexlab{a}}.

\bibitem[Hu et~al.(2024{\natexlab{b}})Hu, Zhao, Qiu, Shu, Hu, Yang, and
  Guo]{hu2024multirc}
Shiyan Hu, Kai Zhao, Xiangfei Qiu, Yang Shu, Jilin Hu, Bin Yang, and Chenjuan
  Guo.
\newblock Multirc: Joint learning for time series anomaly prediction and
  detection with multi-scale reconstructive contrast.
\newblock \emph{arXiv preprint arXiv:2410.15997}, 2024{\natexlab{b}}.

\bibitem[Hu et~al.(2026{\natexlab{a}})Hu, Zhang, Jin, Qiu, Yang, and
  Guo]{huteamWork}
Shiyan Hu, Tengxue Zhang, Jianxin Jin, Xiangfei Qiu, Bin Yang, and Chenjuan
  Guo.
\newblock Teamwork: Multivariate time series anomaly detection via asymmetric
  role-aware channel modeling.
\newblock In \emph{ICML}, 2026{\natexlab{a}}.

\bibitem[Hu et~al.(2025)Hu, Wang, Ding, Wu, Zhang, Li, Wang, Zhang, Li, and
  Chen]{flowts}
Yang Hu, Xiao Wang, Zezhen Ding, Lirong Wu, Huatian Zhang, Stan~Z. Li, Sheng
  Wang, Jiheng Zhang, Ziyun Li, and Tianlong Chen.
\newblock Flowts: Time series generation via rectified flow, 2025.
\newblock URL \url{https://arxiv.org/abs/2411.07506}.

\bibitem[Hu et~al.(2026{\natexlab{b}})Hu, Yang, Zhou, Liu, Tang, Jin, and
  Sun]{hu2026bridging}
Yifan Hu, Jie Yang, Tian Zhou, Peiyuan Liu, Yujin Tang, Rong Jin, and Liang
  Sun.
\newblock Bridging past and future: Distribution-aware alignment for time
  series forecasting.
\newblock In \emph{ICLR}, 2026{\natexlab{b}}.

\bibitem[Huang et~al.(2025{\natexlab{a}})Huang, Zhou, Yang, Yi, Wang, and
  Wang]{huang2025timebase}
Qihe Huang, Zhengyang Zhou, Kuo Yang, Zhongchao Yi, Xu~Wang, and Yang Wang.
\newblock {TimeBase}: The power of minimalism in efficient long-term time
  series forecasting.
\newblock In \emph{International Conference on Machine Learning}, volume 267,
  pp.\  26227--26246. PMLR, 2025{\natexlab{a}}.

\bibitem[Huang et~al.(2025{\natexlab{b}})Huang, Zhao, Li, and Bai]{Timekan}
Songtao Huang, Zhen Zhao, Can Li, and Lei Bai.
\newblock Timekan: Kan-based frequency decomposition learning architecture for
  long-term time series forecasting.
\newblock In \emph{ICLR}, 2025{\natexlab{b}}.

\bibitem[Huang et~al.(2022)Huang, Yang, Wang, Wang, Zhang, Xu, Chen, and
  Vazirgiannis]{huang2022dgraph}
Xuanwen Huang, Yang Yang, Yang Wang, Chunping Wang, Zhisheng Zhang, Jiarong Xu,
  Lei Chen, and Michalis Vazirgiannis.
\newblock Dgraph: {A} large-scale financial dataset for graph anomaly
  detection.
\newblock In \emph{NeurIPS}, pp.\  22765--22777, 2022.

\bibitem[Hyndman et~al.(2008)Hyndman, Koehler, Ord, and
  Snyder]{hyndman2008forecasting}
Rob Hyndman, Anne~B Koehler, J~Keith Ord, and Ralph~D Snyder.
\newblock \emph{Forecasting with exponential smoothing: the state space
  approach}.
\newblock 2008.

\bibitem[Kollovieh et~al.(2025)Kollovieh, Lienen, L{\"{u}}dke, Schwinn, and
  G{\"{u}}nnemann]{TSFlow}
Marcel Kollovieh, Marten Lienen, David L{\"{u}}dke, Leo Schwinn, and Stephan
  G{\"{u}}nnemann.
\newblock Flow matching with gaussian process priors for probabilistic time
  series forecasting.
\newblock In \emph{ICLR}, 2025.

\bibitem[Kulis \& Jordan(2012)Kulis and Jordan]{DBLP:conf/icml/KulisJ12}
Brian Kulis and Michael~I. Jordan.
\newblock Revisiting k-means: New algorithms via bayesian nonparametrics.
\newblock In \emph{ICML}, 2012.

\bibitem[Lago et~al.(2021)Lago, Marcjasz, De~Schutter, and Weron]{EPF}
Jesus Lago, Grzegorz Marcjasz, Bart De~Schutter, and Rafa{\l} Weron.
\newblock Forecasting day-ahead electricity prices: A review of
  state-of-the-art algorithms, best practices and an open-access benchmark.
\newblock \emph{Applied Energy}, 293:\penalty0 116983, 2021.

\bibitem[Li et~al.(2019)Li, Yan, Yang, and Jin]{DSSMF}
Longyuan Li, Junchi Yan, Xiaokang Yang, and Yaohui Jin.
\newblock Learning interpretable deep state space model for probabilistic time
  series forecasting.
\newblock In \emph{IJCAI}, pp.\  2901--2908, 2019.

\bibitem[Li et~al.(2023)Li, Chen, Chen, Wang, Tian, and Zhou]{PUAD}
Yuxin Li, Wenchao Chen, Bo~Chen, Dongsheng Wang, Long Tian, and Mingyuan Zhou.
\newblock Prototype-oriented unsupervised anomaly detection for multivariate
  time series.
\newblock In \emph{ICML}, volume 202, pp.\  19407--19424, 2023.

\bibitem[Li et~al.(2025)Li, Qiu, Chen, Wang, Cheng, Shu, Hu, Guo, Zhou, Wen,
  et~al.]{li2025TSFM-Bench}
Zhe Li, Xiangfei Qiu, Peng Chen, Yihang Wang, Hanyin Cheng, Yang Shu, Jilin Hu,
  Chenjuan Guo, Aoying Zhou, Qingsong Wen, et~al.
\newblock {TSFM-Bench}: A comprehensive and unified benchmark of foundation
  models for time series forecasting.
\newblock In \emph{SIGKDD}, 2025.

\bibitem[Li et~al.(2026)Li, Qiu, Zhu, Wu, Hu, Guo, and Yang]{GCGNet}
Zhengyu Li, Xiangfei Qiu, Yuhan Zhu, Xingjian Wu, Jilin Hu, Chenjuan Guo, and
  Bin Yang.
\newblock {GCGN}et: Graph-consistent generative network for time series
  forecasting with exogenous variables.
\newblock In \emph{ICLR}, 2026.

\bibitem[Lim et~al.(2021)Lim, Ar{\i}k, Loeff, and Pfister]{lim2021temporal}
Bryan Lim, Sercan~{\"O} Ar{\i}k, Nicolas Loeff, and Tomas Pfister.
\newblock Temporal fusion transformers for interpretable multi-horizon time
  series forecasting.
\newblock \emph{International journal of forecasting}, 37\penalty0
  (4):\penalty0 1748--1764, 2021.

\bibitem[Lin et~al.(2024{\natexlab{a}})Lin, Lin, Wu, Wang, Xu, and
  Wang]{lin2024cocv}
Shengsheng Lin, Weiwei Lin, Keyi Wu, Songbo Wang, Minxian Xu, and James~Z Wang.
\newblock Cocv: A compression algorithm for time-series data with continuous
  constant values in iot-based monitoring systems.
\newblock \emph{Internet of Things}, 25:\penalty0 101049, 2024{\natexlab{a}}.

\bibitem[Lin et~al.(2024{\natexlab{b}})Lin, Lin, Wu, Chen, and
  Yang]{lin2024sparsetsf}
Shengsheng Lin, Weiwei Lin, Wentai Wu, Haojun Chen, and Junjie Yang.
\newblock {SparseTSF}: Modeling long-term time series forecasting with 1k
  parameters.
\newblock In \emph{ICML}, pp.\  30211--30226, 2024{\natexlab{b}}.

\bibitem[Lin et~al.(2024{\natexlab{c}})Lin, Lin, Xinyi, Wu, Mo, and
  Zhong]{lincyclenet}
Shengsheng Lin, Weiwei Lin, HU~Xinyi, Wentai Wu, Ruichao Mo, and Haocheng
  Zhong.
\newblock Cyclenet: Enhancing time series forecasting through modeling periodic
  patterns.
\newblock In \emph{NeurIPS}, 2024{\natexlab{c}}.

\bibitem[Liu et~al.(2024{\natexlab{a}})Liu, Yang, Li, and Hong]{RATD}
Jingwei Liu, Ling Yang, Hongyan Li, and Shenda Hong.
\newblock Retrieval-augmented diffusion models for time series forecasting.
\newblock In \emph{NeurIPS}, 2024{\natexlab{a}}.

\bibitem[Liu et~al.(2026{\natexlab{a}})Liu, Qiu, Cheng, Wu, Guo, Yang, and
  Hu]{liu2026astgi}
Xvyuan Liu, Xiangfei Qiu, Hanyin Cheng, Xingjian Wu, Chenjuan Guo, Bin Yang,
  and Jilin Hu.
\newblock Astgi: Adaptive spatio-temporal graph interactions for irregular
  multivariate time series forecasting.
\newblock In \emph{ICLR}, 2026{\natexlab{a}}.

\bibitem[Liu et~al.(2026{\natexlab{b}})Liu, Qiu, Wu, Li, Guo, Hu, and
  Yang]{liu2026rethinking}
Xvyuan Liu, Xiangfei Qiu, Xingjian Wu, Zhengyu Li, Chenjuan Guo, Jilin Hu, and
  Bin Yang.
\newblock Rethinking irregular time series forecasting: A simple yet effective
  baseline.
\newblock In \emph{AAAI}, 2026{\natexlab{b}}.

\bibitem[Liu et~al.(2024{\natexlab{b}})Liu, Hu, Zhang, Wu, Wang, Ma, and
  Long]{iTransformer}
Yong Liu, Tengge Hu, Haoran Zhang, Haixu Wu, Shiyu Wang, Lintao Ma, and
  Mingsheng Long.
\newblock itransformer: Inverted transformers are effective for time series
  forecasting.
\newblock In \emph{ICLR}, 2024{\natexlab{b}}.

\bibitem[Liu et~al.(2025)Liu, Qin, Shi, Chen, Yang, Huang, Wang, and
  Long]{sundial}
Yong Liu, Guo Qin, Zhiyuan Shi, Zhi Chen, Caiyin Yang, Xiangdong Huang, Jianmin
  Wang, and Mingsheng Long.
\newblock Sundial: {A} family of highly capable time series foundation models.
\newblock In \emph{ICML}, 2025.

\bibitem[Lu et~al.(2026{\natexlab{a}})Lu, Chen, Guo, Shu, Wang, and
  Yang]{lu2026dtaf}
Junkai Lu, Peng Chen, Chenjuan Guo, Yang Shu, Meng Wang, and Bin Yang.
\newblock Towards non-stationary time series forecasting with temporal
  stabilization and frequency differencing.
\newblock In \emph{AAAI}, 2026{\natexlab{a}}.

\bibitem[Lu et~al.(2026{\natexlab{b}})Lu, Chen, Wu, Shu, Guo, Jensen, and
  Yang]{lu2026patra}
Junkai Lu, Peng Chen, Xingjian Wu, Yang Shu, Chenjuan Guo, Christian~S. Jensen,
  and Bin Yang.
\newblock {PATRA}: Pattern-aware alignment and balanced reasoning for time
  series question answering.
\newblock In \emph{ICML}, 2026{\natexlab{b}}.

\bibitem[Miao et~al.(2021)Miao, Wu, Wang, Gao, Mao, and
  Yin]{miao2021generative}
Xiaoye Miao, Yangyang Wu, Jun Wang, Yunjun Gao, Xudong Mao, and Jianwei Yin.
\newblock Generative semi-supervised learning for multivariate time series
  imputation.
\newblock In \emph{{AAAI}}, volume~35, pp.\  8983--8991, 2021.

\bibitem[Nie et~al.(2023)Nie, Nguyen, Sinthong, and Kalagnanam]{patchtst}
Yuqi Nie, Nam~H. Nguyen, Phanwadee Sinthong, and Jayant Kalagnanam.
\newblock A time series is worth 64 words: Long-term forecasting with
  transformers.
\newblock In \emph{ICLR}, 2023.

\bibitem[Ning et~al.(2025)Ning, Pan, Liu, Jiang, Zhang, Rasul, Schneider, Ma,
  Nevmyvaka, and Song]{TS-RAG}
Kanghui Ning, Zijie Pan, Yu~Liu, Yushan Jiang, James~Y. Zhang, Kashif Rasul,
  Anderson Schneider, Lintao Ma, Yuriy Nevmyvaka, and Dongjin Song.
\newblock {TS-RAG:} retrieval-augmented generation based time series foundation
  models are stronger zero-shot forecaster.
\newblock In \emph{NeurIPS}, 2025.

\bibitem[Olivares et~al.(2023)Olivares, Challu, Marcjasz, Weron, and
  Dubrawski]{olivares2023neural}
Kin~G Olivares, Cristian Challu, Grzegorz Marcjasz, Rafa{\l} Weron, and Artur
  Dubrawski.
\newblock Neural basis expansion analysis with exogenous variables: Forecasting
  electricity prices with nbeatsx.
\newblock \emph{International Journal of Forecasting}, 39\penalty0
  (2):\penalty0 884--900, 2023.

\bibitem[Pan et~al.(2023)Pan, Wang, Zhang, Yang, Cheng, Chen, Guo, Wen, Tian,
  Dou, et~al.]{pan2023magicscaler}
Zhicheng Pan, Yihang Wang, Yingying Zhang, Sean~Bin Yang, Yunyao Cheng, Peng
  Chen, Chenjuan Guo, Qingsong Wen, Xiduo Tian, Yunliang Dou, et~al.
\newblock Magicscaler: Uncertainty-aware, predictive autoscaling.
\newblock \emph{Proc. {VLDB} Endow.}, 16\penalty0 (12):\penalty0 3808--3821,
  2023.

\bibitem[Paszke et~al.(2019)Paszke, Gross, Massa, Lerer, Bradbury, Chanan,
  Killeen, Lin, Gimelshein, Antiga, Desmaison, K{\"{o}}pf, Yang, DeVito,
  Raison, Tejani, Chilamkurthy, Steiner, Fang, Bai, and
  Chintala]{paszke2019pytorch}
Adam Paszke, Sam Gross, Francisco Massa, Adam Lerer, James Bradbury, Gregory
  Chanan, Trevor Killeen, Zeming Lin, Natalia Gimelshein, Luca Antiga, Alban
  Desmaison, Andreas K{\"{o}}pf, Edward~Z. Yang, Zachary DeVito, Martin Raison,
  Alykhan Tejani, Sasank Chilamkurthy, Benoit Steiner, Lu~Fang, Junjie Bai, and
  Soumith Chintala.
\newblock Pytorch: An imperative style, high-performance deep learning library.
\newblock In \emph{NeurIPS}, pp.\  8024--8035, 2019.

\bibitem[Qiu et~al.(2024)Qiu, Hu, Zhou, Wu, Du, Zhang, Guo, Zhou, Jensen,
  Sheng, and Yang]{qiu2024tfb}
Xiangfei Qiu, Jilin Hu, Lekui Zhou, Xingjian Wu, Junyang Du, Buang Zhang,
  Chenjuan Guo, Aoying Zhou, Christian~S. Jensen, Zhenli Sheng, and Bin Yang.
\newblock {TFB:} towards comprehensive and fair benchmarking of time series
  forecasting methods.
\newblock \emph{Proc. {VLDB} Endow.}, 17\penalty0 (9):\penalty0 2363--2377,
  2024.

\bibitem[Qiu et~al.(2025{\natexlab{a}})Qiu, Li, Qiu, Hu, Zhou, Wu, Li, Guo,
  Zhou, Sheng, Hu, Jensen, and Yang]{qiu2025tab}
Xiangfei Qiu, Zhe Li, Wanghui Qiu, Shiyan Hu, Lekui Zhou, Xingjian Wu, Zhengyu
  Li, Chenjuan Guo, Aoying Zhou, Zhenli Sheng, Jilin Hu, Christian~S. Jensen,
  and Bin Yang.
\newblock {TAB}: Unified benchmarking of time series anomaly detection methods.
\newblock In \emph{Proc. {VLDB} Endow.}, 2025{\natexlab{a}}.

\bibitem[Qiu et~al.(2025{\natexlab{b}})Qiu, Wu, Cheng, Liu, Guo, Hu, and
  Yang]{qiu2025DBLoss}
Xiangfei Qiu, Xingjian Wu, Hanyin Cheng, Xvyuan Liu, Chenjuan Guo, Jilin Hu,
  and Bin Yang.
\newblock {DBLoss}: Decomposition-based loss function for time series
  forecasting.
\newblock In \emph{NeurIPS}, 2025{\natexlab{b}}.

\bibitem[Qiu et~al.(2025{\natexlab{c}})Qiu, Wu, Lin, Guo, Hu, and
  Yang]{qiu2025duet}
Xiangfei Qiu, Xingjian Wu, Yan Lin, Chenjuan Guo, Jilin Hu, and Bin Yang.
\newblock {DUET}: Dual clustering enhanced multivariate time series
  forecasting.
\newblock In \emph{SIGKDD}, pp.\  1185--1196, 2025{\natexlab{c}}.

\bibitem[Qiu et~al.(2026{\natexlab{a}})Qiu, Cheng, Wu, Lu, Hu, Guo, Jensen, and
  Yang]{qiu2026comprehensive}
Xiangfei Qiu, Hanyin Cheng, Xingjian Wu, Junkai Lu, Jilin Hu, Chenjuan Guo,
  Christian~S Jensen, and Bin Yang.
\newblock A comprehensive survey of deep learning for multivariate time series
  forecasting: A channel strategy perspective.
\newblock In \emph{IJCAI}, 2026{\natexlab{a}}.

\bibitem[Qiu et~al.(2026{\natexlab{b}})Qiu, Yan, Liu, Wu, and
  Hu]{qiu2026bridging}
Xiangfei Qiu, Kangjia Yan, Xvyuan Liu, Xingjian Wu, and Jilin Hu.
\newblock Bridging time and frequency: A joint modeling framework for irregular
  multivariate time series forecasting.
\newblock In \emph{ICML}, 2026{\natexlab{b}}.

\bibitem[Qiu et~al.(2026{\natexlab{c}})Qiu, Zhu, Li, Wu, Yang, and
  Hu]{qiu2026dag}
Xiangfei Qiu, Yuhan Zhu, Zhengyu Li, Xingjian Wu, Bin Yang, and Jilin Hu.
\newblock Dag: A dual correlation network for time series forecasting with
  exogenous variables.
\newblock In \emph{ICML}, 2026{\natexlab{c}}.

\bibitem[Rangapuram et~al.(2018)Rangapuram, Seeger, Gasthaus, Stella, Wang, and
  Januschowski]{DeepState}
Syama~Sundar Rangapuram, Matthias~W. Seeger, Jan Gasthaus, Lorenzo Stella,
  Yuyang Wang, and Tim Januschowski.
\newblock Deep state space models for time series forecasting.
\newblock In \emph{NeurIPS}, pp.\  7796--7805, 2018.

\bibitem[Rasul et~al.(2021)Rasul, Seward, Schuster, and Vollgraf]{TimeGrad}
Kashif Rasul, Calvin Seward, Ingmar Schuster, and Roland Vollgraf.
\newblock Autoregressive denoising diffusion models for multivariate
  probabilistic time series forecasting.
\newblock In \emph{ICML}, volume 139, pp.\  8857--8868, 2021.

\bibitem[Salinas et~al.(2020)Salinas, Flunkert, Gasthaus, and
  Januschowski]{DeepAR}
David Salinas, Valentin Flunkert, Jan Gasthaus, and Tim Januschowski.
\newblock Deepar: Probabilistic forecasting with autoregressive recurrent
  networks.
\newblock \emph{International journal of forecasting}, 36\penalty0
  (3):\penalty0 1181--1191, 2020.

\bibitem[Sezer et~al.(2020)Sezer, Gudelek, and
  {\"{O}}zbayoglu]{sezer2020financial}
Omer~Berat Sezer, Mehmet~Ugur Gudelek, and Ahmet~Murat {\"{O}}zbayoglu.
\newblock Financial time series forecasting with deep learning : {A} systematic
  literature review: 2005-2019.
\newblock \emph{Appl. Soft Comput.}, 90:\penalty0 106181, 2020.

\bibitem[Shang \& Chen(2024)Shang and Chen]{MSHyper}
Zongjiang Shang and Ling Chen.
\newblock {MSHyper: Multi-scale hypergraph transformer for long-range time
  series forecasting}.
\newblock \emph{arXiv preprint arXiv:2401.09261}, 2024.

\bibitem[Shang et~al.(2024)Shang, Chen, Wu, and Cui]{AdaMSHyper}
Zongjiang Shang, Ling Chen, Binqing Wu, and Dongliang Cui.
\newblock {Ada-MSHyper: Adaptive multi-scale hypergraph transformer for time
  series forecasting}.
\newblock In \emph{NeurIPS}, volume~37, pp.\  33310--33337, 2024.

\bibitem[Shannon(1948)]{ShannonEntropy}
Claude~E. Shannon.
\newblock A mathematical theory of communication.
\newblock \emph{Bell Syst. Tech. J.}, 27\penalty0 (3):\penalty0 379--423, 1948.

\bibitem[Shao et~al.(2025)Shao, Fang, Chen, and Gao]{FOTraj}
Wei Shao, Ziquan Fang, Lu~Chen, and Yunjun Gao.
\newblock Towards trajectory anomaly detection: a fine-grained and
  noise-resilient framework.
\newblock In \emph{SIGKDD}, pp.\  2490--2501, 2025.

\bibitem[Shen(2025)]{H-PAD}
Ke{-}Yuan Shen.
\newblock Learn hybrid prototypes for multivariate time series anomaly
  detection.
\newblock In \emph{ICLR}, 2025.

\bibitem[Song et~al.(2023)Song, Kim, Oh, and Cho]{memto}
Junho Song, Keonwoo Kim, Jeonglyul Oh, and Sungzoon Cho.
\newblock {MEMTO}: Memory-guided transformer for multivariate time series
  anomaly detection.
\newblock In \emph{NeurIPS 2023}, 2023.

\bibitem[Stitsyuk \& Choi(2025)Stitsyuk and Choi]{xPatch}
Artyom Stitsyuk and Jaesik Choi.
\newblock xpatch: Dual-stream time series forecasting with exponential
  seasonal-trend decomposition.
\newblock In \emph{AAAI}, volume~39, pp.\  20601--20609, 2025.

\bibitem[Tashiro et~al.(2021)Tashiro, Song, Song, and Ermon]{CSDI}
Yusuke Tashiro, Jiaming Song, Yang Song, and Stefano Ermon.
\newblock {CSDI:} conditional score-based diffusion models for probabilistic
  time series imputation.
\newblock In \emph{NeurIPS}, pp.\  24804--24816, 2021.

\bibitem[Tayal et~al.(2024)Tayal, Renganathan, Jia, Kumar, and
  Lu]{tayal2024exotst}
Kshitij Tayal, Arvind Renganathan, Xiaowei Jia, Vipin Kumar, and Dan Lu.
\newblock Exotst: Exogenous-aware temporal sequence transformer for time series
  prediction.
\newblock In \emph{ICDM}, pp.\  857--862, 2024.

\bibitem[Vagropoulos et~al.(2016)Vagropoulos, Chouliaras, Kardakos, Simoglou,
  and Bakirtzis]{vagropoulos2016comparison}
Stylianos~I Vagropoulos, GI~Chouliaras, Evaggelos~G Kardakos, Christos~K
  Simoglou, and Anastasios~G Bakirtzis.
\newblock Comparison of sarimax, sarima, modified sarima and ann-based models
  for short-term pv generation forecasting.
\newblock In \emph{ENERGYCON}, pp.\  1--6, 2016.

\bibitem[Vandermeulen et~al.(2024)Vandermeulen, Tai, and
  Aragam]{DBLP:conf/nips/VandermeulenTA24}
Robert~A. Vandermeulen, Wai~Ming Tai, and Bryon Aragam.
\newblock Breaking the curse of dimensionality in structured density
  estimation.
\newblock In \emph{NeurIPS}, 2024.

\bibitem[Wang et~al.(2025{\natexlab{a}})Wang, Li, Chen, Chen, Dai, Wang, Li,
  and Lin]{wang2026time}
Hao Wang, Pan Li, Zhichao Chen, Xu~Chen, Qingyang Dai, Lei Wang, Haoxuan Li,
  and Zhouchen Lin.
\newblock Time-o1: Time-series forecasting needs transformed label alignment.
\newblock In \emph{NeurIPS}, 2025{\natexlab{a}}.

\bibitem[Wang et~al.(2025{\natexlab{b}})Wang, Pan, Shen, Chen, Yang, Yang,
  Zhang, Liu, Li, and Tao]{wang2025iclrfredf}
Hao Wang, Licheng Pan, Yuan Shen, Zhichao Chen, Degui Yang, Yifei Yang, Sen
  Zhang, Xinggao Liu, Haoxuan Li, and Dacheng Tao.
\newblock Fredf: learning to forecast in the frequency domain.
\newblock In \emph{ICLR}, 2025{\natexlab{b}}.

\bibitem[Wang et~al.(2026)Wang, Pan, Lu, Chu, Li, He, Chen, Li, Wen, and
  Lin]{wang2026iclrdistdf}
Hao Wang, Licheng Pan, Yuan Lu, Zhixuan Chu, Xiaoxi Li, Shuting He, Zhichao
  Chen, Haoxuan Li, Qingsong Wen, and Zhouchen Lin.
\newblock Distdf: time-series forecasting needs joint-distribution wasserstein
  alignment.
\newblock In \emph{ICLR}, 2026.

\bibitem[Wang et~al.(2024)Wang, Wu, Dong, Qin, Zhang, Liu, Qiu, Wang, and
  Long]{wang2024timexer}
Yuxuan Wang, Haixu Wu, Jiaxiang Dong, Guo Qin, Haoran Zhang, Yong Liu, Yunzhong
  Qiu, Jianmin Wang, and Mingsheng Long.
\newblock Timexer: Empowering transformers for time series forecasting with
  exogenous variables.
\newblock In \emph{NeurIPS}, volume~37, pp.\  469--498, 2024.

\bibitem[Williams(2001)]{williams2001multivariate}
Billy~M Williams.
\newblock Multivariate vehicular traffic flow prediction: Evaluation of arimax
  modeling.
\newblock \emph{Transportation Research Record}, 1776\penalty0 (1):\penalty0
  194--200, 2001.

\bibitem[Wu et~al.(2025{\natexlab{a}})Wu, Shang, Huang, and
  Chen]{wu2025millgnn}
Binqing Wu, Zongjiang Shang, Jianlong Huang, and Ling Chen.
\newblock Millgnn: Learning multi-scale lead-lag dependencies for multi-variate
  time series forecasting.
\newblock In \emph{CIKM}, pp.\  3344--3354, 2025{\natexlab{a}}.

\bibitem[Wu et~al.(2021)Wu, Xu, Wang, and Long]{Autoformer}
Haixu Wu, Jiehui Xu, Jianmin Wang, and Mingsheng Long.
\newblock Autoformer: Decomposition transformers with auto-correlation for
  long-term series forecasting.
\newblock In \emph{NeurIPS}, pp.\  22419--22430, 2021.

\bibitem[Wu et~al.(2025{\natexlab{b}})Wu, Qiu, Cheng, Li, Hu, Guo, and
  Yang]{wu2025srsnet}
Xingjian Wu, Xiangfei Qiu, Hanyin Cheng, Zhengyu Li, Jilin Hu, Chenjuan Guo,
  and Bin Yang.
\newblock Enhancing time series forecasting through selective representation
  spaces: A patch perspective.
\newblock In \emph{NeurIPS}, 2025{\natexlab{b}}.

\bibitem[Wu et~al.(2025{\natexlab{c}})Wu, Qiu, Gao, Hu, Yang, and
  Guo]{wu2025k2vae}
Xingjian Wu, Xiangfei Qiu, Hongfan Gao, Jilin Hu, Bin Yang, and Chenjuan Guo.
\newblock {K${}^2$VAE}: A koopman-kalman enhanced variational autoencoder for
  probabilistic time series forecasting.
\newblock In \emph{ICML}, 2025{\natexlab{c}}.

\bibitem[Wu et~al.(2025{\natexlab{d}})Wu, Qiu, Li, Wang, Hu, Guo, Xiong, and
  Yang]{wu2025catch}
Xingjian Wu, Xiangfei Qiu, Zhengyu Li, Yihang Wang, Jilin Hu, Chenjuan Guo, Hui
  Xiong, and Bin Yang.
\newblock {CATCH}: Channel-aware multivariate time series anomaly detection via
  frequency patching.
\newblock In \emph{ICLR}, 2025{\natexlab{d}}.

\bibitem[Wu et~al.(2026)Wu, Lu, Li, Qiu, Hu, Guo, Jensen, and
  Yang]{wu2026timeart}
Xingjian Wu, Junkai Lu, Zhengyu Li, Xiangfei Qiu, Jilin Hu, Chenjuan Guo,
  Christian~S Jensen, and Bin Yang.
\newblock {TimeART}: Towards agentic time series reasoning via
  tool-augmentation.
\newblock \emph{arXiv preprint arXiv:2601.13653}, 2026.

\bibitem[Wu et~al.(2024)Wu, Wu, Yang, Zhou, Guo, Qiu, Hu, Sheng, and
  Jensen]{wu2024autocts++}
Xinle Wu, Xingjian Wu, Bin Yang, Lekui Zhou, Chenjuan Guo, Xiangfei Qiu, Jilin
  Hu, Zhenli Sheng, and Christian~S Jensen.
\newblock {AutoCTS++}: zero-shot joint neural architecture and hyperparameter
  search for correlated time series forecasting.
\newblock \emph{The VLDB Journal}, 33\penalty0 (5):\penalty0 1743--1770, 2024.

\bibitem[Yang et~al.(2025{\natexlab{a}})Yang, Hu, Zhang, Niu, Yu, and
  Ding]{yang2025revisiting}
Jie Yang, Yifan Hu, Kexin Zhang, Luyang Niu, Philip~S Yu, and Kaize Ding.
\newblock Revisiting multivariate time series forecasting with missing values.
\newblock \emph{arXiv preprint arXiv:2509.23494}, 2025{\natexlab{a}}.

\bibitem[Yang et~al.(2025{\natexlab{b}})Yang, Zhang, Zhang, Yu, and
  Ding]{yang2025glocal}
Jie Yang, Kexin Zhang, Guibin Zhang, Philip~S Yu, and Kaize Ding.
\newblock Glocal information bottleneck for time series imputation.
\newblock \emph{arXiv preprint arXiv:2510.04910}, 2025{\natexlab{b}}.

\bibitem[Yang \& van Leeuwen(2024)Yang and van Leeuwen]{DBLP:conf/nips/YangL24}
Lincen Yang and Matthijs van Leeuwen.
\newblock Conditional density estimation with histogram trees.
\newblock In \emph{NeurIPS}, 2024.

\bibitem[Yang et~al.(2026)Yang, Jiang, Xu, Zhou, Yang, Zhu, Geng, Zeng, Chen,
  Gu, Jin, and Sun]{Baguan-TS}
Linxiao Yang, Xue Jiang, Gezheng Xu, Tian Zhou, Min Yang, Zhaoyang Zhu, Linyuan
  Geng, Zhipeng Zeng, Qiming Chen, Xinyue Gu, Rong Jin, and Liang Sun.
\newblock Baguan-ts: {A} sequence-native in-context learning model for time
  series forecasting with covariates.
\newblock \emph{CoRR}, abs/2603.17439, 2026.

\bibitem[Yu et~al.(2025{\natexlab{a}})Yu, Wang, Shao, Qian, Zhang, Wei, An,
  Wang, and Xu]{11002729}
Chengqing Yu, Fei Wang, Zezhi Shao, Tangwen Qian, Zhao Zhang, Wei Wei, Zhulin
  An, Qi~Wang, and Yongjun Xu.
\newblock Ginar+: A robust end-to-end framework for multivariate time series
  forecasting with missing values.
\newblock \emph{IEEE Transactions on Knowledge and Data Engineering},
  37\penalty0 (8):\penalty0 4635--4648, 2025{\natexlab{a}}.

\bibitem[Yu et~al.(2025{\natexlab{b}})Yu, Wang, Yang, Shao, Sun, Qian, Wei, An,
  and Xu]{yu2025merlin}
Chengqing Yu, Fei Wang, Chuanguang Yang, Zezhi Shao, Tao Sun, Tangwen Qian, Wei
  Wei, Zhulin An, and Yongjun Xu.
\newblock Merlin: Multi-view representation learning for robust multivariate
  time series forecasting with unfixed missing rates.
\newblock In \emph{SIGKDD}, pp.\  3633--3644, 2025{\natexlab{b}}.

\bibitem[Zeng et~al.(2023)Zeng, Chen, Zhang, and Xu]{Zengdlinear}
Ailing Zeng, Muxi Chen, Lei Zhang, and Qiang Xu.
\newblock Are transformers effective for time series forecasting?
\newblock In \emph{{AAAI}}, volume~37, pp.\  11121--11128, 2023.

\bibitem[Zhang et~al.(2024)Zhang, Wen, Zhang, Zheng, Li, and Bian]{ProbTS}
Jiawen Zhang, Xumeng Wen, Zhenwei Zhang, Shun Zheng, Jia Li, and Jiang Bian.
\newblock {ProbTS}: Benchmarking point and distributional forecasting across
  diverse prediction horizons.
\newblock In \emph{NeurIPS}, 2024.

\bibitem[Zhou et~al.(2021)Zhou, Zhang, Peng, Zhang, Li, Xiong, and
  Zhang]{Informer}
Haoyi Zhou, Shanghang Zhang, Jieqi Peng, Shuai Zhang, Jianxin Li, Hui Xiong,
  and Wancai Zhang.
\newblock Informer: Beyond efficient transformer for long sequence time-series
  forecasting.
\newblock In \emph{{AAAI}}, volume~35, pp.\  11106--11115, 2021.

\bibitem[Zhou et~al.(2025)Zhou, Liu, Liang, Song, and Li]{zhou2025crosslinear}
Pengfei Zhou, Yunlong Liu, Junli Liang, Qi~Song, and Xiangyang Li.
\newblock Crosslinear: Plug-and-play cross-correlation embedding for time
  series forecasting with exogenous variables.
\newblock In \emph{SIGKDD}, 2025.

\bibitem[Zhou et~al.(2022)Zhou, Ma, Wen, Wang, Sun, and Jin]{zhou2022fedformer}
Tian Zhou, Ziqing Ma, Qingsong Wen, Xue Wang, Liang Sun, and Rong Jin.
\newblock Fedformer: Frequency enhanced decomposed transformer for long-term
  series forecasting.
\newblock In \emph{{ICML}}, pp.\  27268--27286, 2022.

\bibitem[Zhu et~al.(2026)Zhu, Hu, Cai, Li, Ma, Qiu, Li, Zhang, Fu, Jiang, and
  Yang]{zhu2026wpbench}
Yuhan Zhu, Jilin Hu, Xinying Cai, Yingshan Li, Li~Ma, Xiangfei Qiu, Linsen Li,
  Kai Zhang, Yao Fu, Weihao Jiang, and Bin Yang.
\newblock {WPBench}: A comprehensive benchmark for wind power forecasting.
\newblock \emph{arXiv preprint arXiv:2609.24444}, 2026.

\end{thebibliography}
\bibliographystyle{paper-style}

\clearpage
\appendix
\section{Experimental Details}
\label{Experimental Details}

\subsection{Datasets}
\label{appendix DATASETS}

\begingroup
\setlength{\intextsep}{6pt}
\begin{table}[H]
\caption{Statistics of datasets. \#Num denotes the number of exogenous variables. Ex. and En. are abbreviations for the exogenous variable and endogenous variable, respectively.}
\label{Multivariate datasets}
\centering
\renewcommand{\arraystretch}{1.15}
\resizebox{\textwidth}{!}{
\begin{tabular}{@{}ccccccl@{}}
\toprule
Dataset & \#Num & Ex. Descriptions & En. Descriptions & Sampling Frequency & Lengths & Split \\ \midrule
NP & 2 & Grid load, wind power & Nord Pool electricity price & 1 Hour & 52,416 & 7:1:2 \\
PJM & 2 & System load, COMED zonal load & COMED zonal electricity price & 1 Hour & 52,416 & 7:1:2 \\
BE & 2 & Belgian load, French generation & Belgian electricity price & 1 Hour & 52,416 & 7:1:2 \\
FR & 2 & Generation, load & French electricity price & 1 Hour & 52,416 & 7:1:2 \\
DE & 2 & Wind power, Amprion zonal load & German electricity price & 1 Hour & 52,416 & 7:1:2 \\
Energy & 5 & Battery, geothermal, hydroelectric, solar, wind & Thermoelectric generation & 1 Hour & 13,064 & 7:1:2 \\
Colbun & 2 & Precipitation, tributary inflow & Water level & 1 Day & 2,958 & 7:1:2 \\
Rapel & 2 & Precipitation, tributary inflow & Water level & 1 Day & 3,366 & 7:1:2 \\
Sdwpfh & 6 & Climate Feature & Active power & 1 Hour & 14,641 & 7:1:2 \\
Sdwpfm & 6 & Climate Feature & Active power & 30 Minutes & 29,281 & 7:1:2 \\
\bottomrule
\end{tabular}}
\end{table}
\endgroup

We conduct experiments on 12 real-world datasets with exogenous variables, including five electricity price datasets from the EPF benchmark~\citep{EPF,wang2024timexer} and seven energy and hydropower datasets collected by DAG~\citep{qiu2026dag}, following KITE~\citep{cheng2026kite} and GCGNet~\citep{GCGNet}.
We evaluate forecasting under the setting where future exogenous variables are provided as inputs.
\textbf{NP} contains hourly electricity prices from the Nord Pool market, with grid-load and wind-power forecasts as exogenous variables.
\textbf{PJM} records zonal electricity prices in the Commonwealth Edison (COMED) area of the Pennsylvania--New Jersey--Maryland Interconnection, with system-load and COMED zonal-load forecasts as exogenous variables.
\textbf{BE} and \textbf{FR} contain Belgian and French electricity prices, respectively; BE uses Belgian load and French generation forecasts as covariates, while FR uses French generation and load forecasts.
\textbf{DE} records hourly German electricity prices, with Amprion zonal-load forecasts and wind-power forecasts as exogenous variables.
\textbf{Energy} provides hourly power-generation data from Chile, where thermoelectric generation is the endogenous target and battery, geothermal, hydroelectric, solar, and wind generation are exogenous variables.
\textbf{Colbun} and \textbf{Rapel} contain daily measurements from two Chilean reservoirs, with water level as the target and precipitation and tributary inflow as covariates.
\textbf{Sdwpfh1}, \textbf{Sdwpfh2}, \textbf{Sdwpfm1}, and \textbf{Sdwpfm2} record wind-power generation from two turbines at the Longyuan wind farm at hourly and half-hourly resolutions, respectively.
Their target is active power output (Patv), and their exogenous variables are six meteorological measurements from ERA5~\citep{era5}: temperature, surface pressure, relative humidity, wind speed, wind direction, and total precipitation.
All datasets are split chronologically into training, validation, and test sets in a $7{:}1{:}2$ ratio, with detailed statistics provided in Table~\ref{Multivariate datasets}.

\subsection{Baselines}
\label{app:baselines}

We compare XMatch with 10 competitive baselines for  forecasting. Following the treatment of future exogenous variables, we organize these methods into two groups.

\textbf{Methods with native future-covariate support.} This group includes DAG~\citep{qiu2026dag}, KITE~\citep{cheng2026kite}, GCGNet~\citep{GCGNet}, TimeXer~\citep{wang2024timexer}, TFT~\citep{lim2021temporal}, and TiDE~\citep{das2023tide}. These methods incorporate future exogenous information within their forecasting architectures. They cover several approaches to covariate modeling, including temporal and cross-variable correlation modeling, graph-based interactions, attention mechanisms, and MLP-based fusion. KITE additionally uses exogenous information to guide generative forecasting; here, we evaluate its point forecasts using the same deterministic metrics as the other methods.

\textbf{Methods augmented with future-covariate fusion.} This group consists of DUET~\citep{qiu2025duet}, CrossLinear~\citep{zhou2025crosslinear}, Amplifier~\citep{amplifier}, and TimeKAN~\citep{Timekan}. To incorporate known future exogenous variables, we equip these baselines with the MLP fusion module described in Appendix~\ref{appendix MLP}, following prior covariate forecasting studies~\citep{qiu2026dag}. The module combines each backbone's output with future exogenous inputs to produce the final endogenous forecast. All baselines are evaluated within the TFB framework~\citep{qiu2024tfb}, using the same dataset splits and forecasting horizons as XMatch.

The code repositories for the baseline models are listed in Table~\ref{tab:baseline-repositories}.

\begin{table}[!ht]
\centering
\caption{Code repositories for the baseline models.}
\label{tab:baseline-repositories}
\begingroup
\footnotesize
\setlength{\tabcolsep}{4pt}
\renewcommand{\arraystretch}{1.2}
\begin{tabularx}{\textwidth}{@{}l>{\raggedright\arraybackslash}X l>{\raggedright\arraybackslash}X@{}}
\toprule
Model & Code repository & Model & Code repository \\
\midrule
DAG & \href{https://github.com/decisionintelligence/DAG}{\textcolor{blue}{decisionintelligence/DAG}} & TiDE & \href{https://github.com/thuml/Time-Series-Library}{\textcolor{blue}{thuml/Time-Series-Library}} \\
KITE & \href{https://github.com/decisionintelligence/KITE}{\textcolor{blue}{decisionintelligence/KITE}} & DUET & \href{https://github.com/decisionintelligence/DUET}{\textcolor{blue}{decisionintelligence/DUET}} \\
GCGNet & \href{https://github.com/decisionintelligence/GCGNet}{\textcolor{blue}{decisionintelligence/GCGNet}} & CrossLinear & \href{https://github.com/mumiao2000/CrossLinear}{\textcolor{blue}{mumiao2000/CrossLinear}} \\
TimeXer & \href{https://github.com/thuml/TimeXer}{\textcolor{blue}{thuml/TimeXer}} & Amplifier & \href{https://github.com/aikunyi/amplifier}{\textcolor{blue}{aikunyi/amplifier}} \\
TFT & \href{https://github.com/google-research/google-research/tree/master/tft}{\textcolor{blue}{google-research/\ldots/tft}} & TimeKAN & \href{https://github.com/huangst21/TimeKAN}{\textcolor{blue}{huangst21/TimeKAN}} \\
\bottomrule
\end{tabularx}
\endgroup
\end{table}

\subsection{Evaluation Metrics}
\label{app:evaluation-metrics}
To evaluate model performance, we employ Mean Absolute Error (MAE) and Mean Squared Error (MSE) for deterministic forecasting, following the TFB benchmark~\cite{qiu2024tfb}.

\textbf{Mean Squared Error (MSE).}
The Mean Squared Error (MSE) measures the average of the squares of the errors—that is, the average squared difference between the estimated values and the actual observation. Unlike MAE, MSE penalizes larger errors more severely due to the squaring operation, making it more sensitive to outliers. The mathematical representation of MSE is defined as:
\begin{align}
\textrm{MSE} = \frac{1}{K \times T} \sum^K_{k=1} \sum^T_{t=1} (x^k_{t} - \hat{x}^k_{t})^2,
\end{align}
where $x^k_{t}$ denotes the ground truth and $\hat{x}^k_{t}$ denotes the predicted value.

\textbf{Mean Absolute Error (MAE).}
The Mean Absolute Error (MAE) measures the average magnitude of the errors in a set of predictions, without considering their direction. It is the average over the verification sample of the absolute differences between prediction and actual observation. The mathematical representation of MAE is given by:
\begin{align}
\textrm{MAE} = \frac{1}{K \times T} \sum^K_{k=1} \sum^T_{t=1} |x^k_{t} - \hat{x}^k_{t}|,
\end{align}
where $K$ represents the number of time series and $T$ represents the number of time steps.

\subsection{Implementation Details}
\label{app:implementation-details}

All experiments are conducted using PyTorch~\citep{paszke2019pytorch} in Python 3.10 and executed on an NVIDIA GeForce RTX 3090 GPU. We do not apply the ``Drop Last'' trick to ensure a fair comparison, retaining the final incomplete batch during evaluation. Testing uses the rolling batch interface of TFB~\citep{qiu2024tfb} to process all test windows, including the final batch when it contains fewer windows than the batch size. The evaluation code additionally verifies that the number of predictions matches the number of target windows.

\subsection{Matching Precision and Historical Support Evaluation}
\label{app:matching-analysis}

We compare direct matching with \textit{ProtoTree} on Energy and Sdwpfh1 using a history length of 168, a forecast length of 24, and a patch length of $P=12$. Direct matching uses increasing numbers of exogenous variables, from one to five on Energy and one to six on Sdwpfh1, while \emph{Tree} denotes the full soft-matching result. Both methods use training patches as historical references and are evaluated by weighted DTW-based similarity to the ground-truth future endogenous pattern and effective historical support. The boxplots summarize these metrics over successfully matched queries, with effective support shown on a logarithmic scale.

\textbf{Weighted DTW-based endogenous pattern similarity.} Let $\widetilde{\boldsymbol{y}}_q\in\mathbb{R}^{P}$ denote the standardized ground-truth future endogenous patch corresponding to query $q$, and let $\widetilde{\boldsymbol{c}}_r$ denote the standardized retrieved endogenous pattern center $r$. The ground-truth future endogenous patch is used only as an evaluation reference and is not used to construct the retrieval query or determine retrieval weights. We compute its DTW-based similarity to each retrieved endogenous pattern and take the weighted average under the retrieved pattern distribution $\pi_q$:
\begin{align}
K_{q,r}&=1-\frac{\operatorname{DTW}_{\mathrm{band}=2}(\widetilde{\boldsymbol{y}}_q,\widetilde{\boldsymbol{c}}_r)}{2P}, \qquad
S_q=\sum_r\pi_{q,r}K_{q,r}.
\end{align}
The DTW band permits temporal offsets of at most two time steps. Higher scores indicate that the retrieved endogenous patterns more closely resemble the ground-truth future endogenous pattern for the query. Direct matching assigns equal weight to every matched historical patch, so $\pi_{q,r}$ is the empirical frequency of endogenous pattern $r$ among the matches for query $q$. For brevity, we omit the query subscript in the aggregation weights and pattern distribution below.

For Tree, we retain the complete aggregation weights $w_v$ from the \textit{ProtoTree Matcher}, excluding the virtual root, and normalize them for this analysis:
\begin{align}
\alpha_v&=\frac{w_v}{\sum_{u\ne v_{\mathrm{root}}}w_u}, \qquad
\pi_r=\sum_{v\ne v_{\mathrm{root}}}\alpha_v p_v(r).
\end{align}
A Tree query is included when its total non-root weight is positive. We do not select a maximum-weight node or truncate to top-$k$ nodes. Here, $p_v(r)$ is the empirical endogenous pattern distribution at node $v$. Thus, the similarity score can equivalently be written as $S_q=\sum_{v\ne v_{\mathrm{root}}}\alpha_v\sum_r p_v(r)K_{q,r}$, incorporating both node aggregation weights and within-node pattern frequencies. Applying the same $S_q$ to both methods evaluates the agreement of their retrieved shape evidence with the ground-truth future endogenous pattern before $\operatorname{MLP}_V$, rather than the smoothness of an averaged response curve.

\textbf{Effective support.} Let $\mathcal{M}_v$ be the set of historical patches supporting node $v$ and $n_v=|\mathcal{M}_v|$. The normalized contribution of historical patch $i$ and the effective support are
\begin{align}
\beta_i&=\sum_{v:\,i\in\mathcal{M}_v}\frac{\alpha_v}{n_v}, \qquad
N_{\mathrm{eff}}=\frac{1}{\sum_i\beta_i^2}.
\end{align}
Contributions of the same patch are combined across nodes before squaring, accounting for overlapping support between ancestors and descendants. For direct matching with $n$ equally weighted historical patches, this definition reduces exactly to $N_{\mathrm{eff}}=n$. Effective support therefore measures the amount of historical evidence under the aggregation weights, rather than the sum of node counts.

\subsection{MLP Fusion Module}
\label{appendix MLP}
For a fair comparison, we equip conventional forecasting methods with an efficient MLP fusion module~\citep{qiu2026dag}, allowing them to incorporate future exogenous information.
Given the historical endogenous variables $X^{\text{endo}} \in \mathbb{R}^{N \times T}$ and historical exogenous variables $X^{\text{exo}}\in \mathbb{R}^{D \times T}$, the forecasting backbone first produces a latent representation:
\begin{equation}
z = \theta_{\text{Model}}(X^{\text{endo}}, X^{\text{exo}}),
\end{equation}
where $z\in \mathbb{R}^{N \times F}$ and $\theta_{\text{Model}}$ denotes the parameters of the backbone. The resulting representation $z$ is then concatenated with the future exogenous variables $Y^{\text{exo}}\in \mathbb{R}^{D \times F}$ and fed into an MLP to obtain the final forecast:
\begin{equation}
\hat{Y}^{\text{endo}} = \theta_{\text{MLP}}\big(\text{Concat}(z, Y^{\text{exo}})\big),
\end{equation}
where $\theta_{\text{MLP}}$ denotes the parameters of the MLP fusion module.

\clearpage
\section{More Analysis}
\label{app:more-analysis}

\subsection{Additional Parameter Sensitivity Analysis}
\label{app:parameter-sensitivity}

\begin{figure}[!ht]
    \centering
    \begin{subfigure}[t]{0.24\textwidth}
        \centering
        \includegraphics[width=\linewidth]{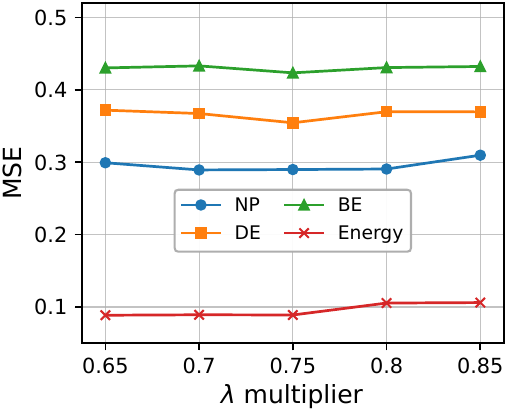}
        \caption{Clustering penalty}
        \label{fig:sensitivity-clustering-penalty}
    \end{subfigure}\hfill
    \begin{subfigure}[t]{0.24\textwidth}
        \centering
        \includegraphics[width=\linewidth]{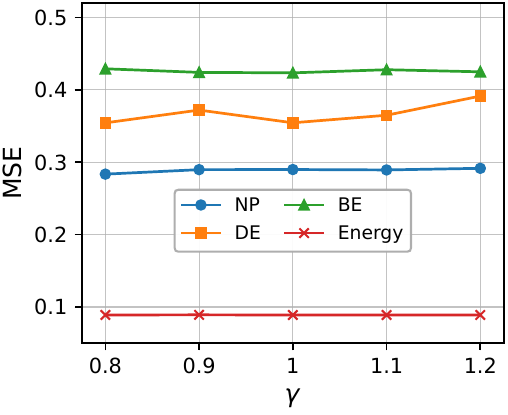}
        \caption{Soft-DTW smoothing}
        \label{fig:sensitivity-softdtw-gamma}
    \end{subfigure}\hfill
    \begin{subfigure}[t]{0.24\textwidth}
        \centering
        \includegraphics[width=\linewidth]{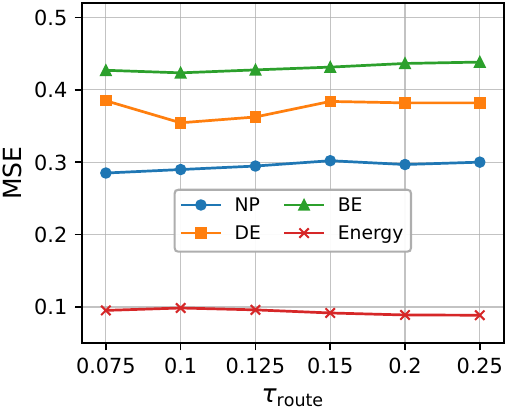}
        \caption{Routing temp.}
        \label{fig:sensitivity-routing-temperature}
    \end{subfigure}\hfill
    \begin{subfigure}[t]{0.24\textwidth}
        \centering
        \includegraphics[width=\linewidth]{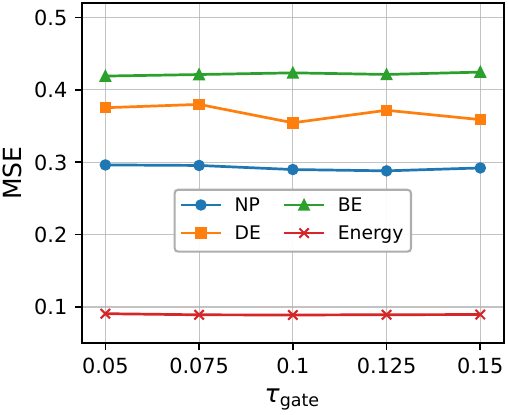}
        \caption{Stopping gate temp.}
        \label{fig:sensitivity-stopping-gate-temperature}
    \end{subfigure}
    \caption{Additional hyperparameter sensitivity of XMatch.}
    \label{fig:additional-parameter-sensitivity}
\end{figure}

We further study the parameter sensitivity of XMatch, including the clustering penalty multiplier and Soft-DTW smoothing parameter $\gamma$ in the \textit{ProtoTree Creator}, and the routing and stopping gate temperatures in the \textit{ProtoTree Matcher}. 1) Figure~\ref{fig:sensitivity-clustering-penalty} shows the impact of the clustering penalty. The multiplier scales the 90th percentile of Soft-DTW divergences (normalized by $P^2$) between each variable's normalized training patches and their mean patch to initialize the clustering penalty. Although performance is relatively stable overall, forecasting errors increase slightly toward the lower or upper ends of the tested range in some cases. A small penalty may produce overly fragmented clusters, whereas an excessively large penalty favors fewer, coarser pattern centers that may obscure useful pattern distinctions. A moderate penalty is therefore preferable for preserving distinct patterns while retaining sufficient samples in each cluster. We recommend selecting the penalty multiplier within $[0.7, 0.75]$. 2) Figure~\ref{fig:sensitivity-softdtw-gamma} illustrates the impact of $\gamma$. Performance remains relatively stable across much of the tested range, but larger values increase forecasting errors on some datasets. This suggests that excessive smoothing across temporal alignments may weaken the distinction between different patch shapes. We recommend selecting $\gamma$ within $[0.8, 1.0]$ to allow flexible temporal alignment while preserving discriminative shape information. 3) As illustrated in Figure~\ref{fig:sensitivity-routing-temperature}, the routing temperature modulates the distribution of aggregation weights across branches, with moderate performance variations as the weights become more diffuse. We recommend $[0.075, 0.125]$ as an initial tuning range, with higher temperatures considered when more diffuse aggregation improves validation performance. 4) Figure~\ref{fig:sensitivity-stopping-gate-temperature} indicates low sensitivity to the stopping gate temperature, which controls the smoothness of the continue-or-stop transition during training; inference still uses a hard threshold. We recommend $0.1$ as a starting value, with tuning within $[0.05, 0.15]$ if needed. These recommendations serve as practical starting points, with final values selected on the validation set.

\subsection{Limitations and Future Work}
\label{app:limitations}

Despite the promising performance of XMatch, this work still has several limitations.

\textbf{1) Fixed Variable Ordering.} The current framework determines a fixed exogenous variable order based on the information gain between each individual exogenous variable and endogenous patterns. Although this ordering prioritizes informative variables for hierarchical matching, individual information gain may not fully reflect the joint contributions of multiple variables, and the most informative variable combinations may vary across forecasting contexts. Exploring tree construction based on conditional information gain, or dynamically adapting variable selection and matching order to each query, is a direction for future work.

\textbf{2) Fixed Patch Length.} The current framework uses a fixed patch length to construct pattern prototypes. While this design enables aligned pattern extraction and matching, a single temporal scale may not fully capture exogenous--endogenous associations underlying both short-term fluctuations and long-term changes. Developing a multi-scale \textit{ProtoTree} or adaptively selecting patch lengths based on the data is another direction for future work.

\clearpage

\subsection{Temporal Recurrence of Exogenous and Endogenous Patterns}
\label{app:temporal-recurrence}

\begin{figure}[!ht]
    \centering
    \includegraphics[width=\linewidth]{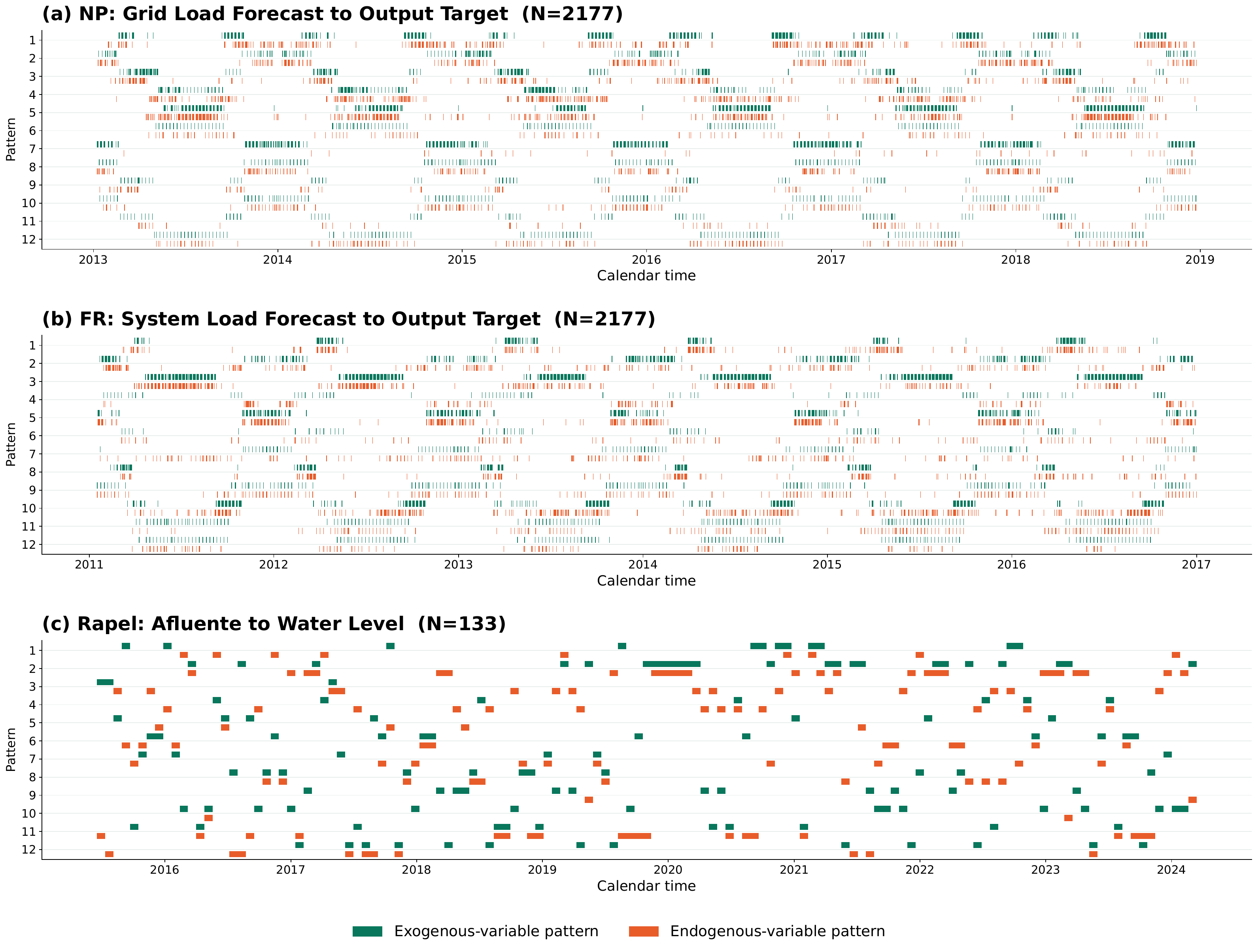}
    \caption{\textbf{Temporal recurrence of exogenous and endogenous patterns} on (a) NP, (b) FR, and (c) Rapel. The horizontal axis shows calendar time, and the vertical axis indexes patterns. Green and orange marks indicate the occurrence intervals of exogenous and endogenous patterns.}
    \label{fig:temporal-recurrence}
\end{figure}

Figure~\ref{fig:temporal-recurrence} complements the pattern correspondences in Figure~\ref{fig:intro} by showing when exogenous and endogenous patterns occur. Both types of patterns recur at multiple, separated times. 
Together, the two figures motivate XMatch from complementary perspectives: Figure~\ref{fig:intro} shows that an exogenous pattern is often associated with a small set of endogenous patterns, while Figure~\ref{fig:temporal-recurrence} shows that both types of patterns recur over time. This suggests that their correspondences from the training set may remain useful in forecasting when similar exogenous patterns reappear. Given future exogenous patterns, XMatch can therefore match them against historical exogenous patterns and use the corresponding endogenous patterns to narrow down the plausible future responses and guide prediction.

\section{Related Works}
\label{app:related-works}

Time series analysis encompasses forecasting~\citep{qiu2024tfb,ProbTS,li2025TSFM-Bench}, anomaly detection~\citep{qiu2025tab}, imputation~\citep{CSDI,miao2021generative,yang2025glocal}, generation~\citep{flowts}, and tool-augmented reasoning~\citep{wu2026timeart}. These tasks share the need to represent temporal regularities. Complementing the main text, we review broader forecasting settings and temporal representation learning.

\subsection{Time Series Forecasting}
\label{app:univariate-multivariate}

\textbf{Univariate forecasting.} Univariate forecasting predicts a single variable from its history. Classical statistical approaches model temporal dependence through autoregression and exponential smoothing~\citep{box1970distribution,hyndman2008forecasting}. Deep forecasting methods learn temporal patterns from data, including recurrent and state-space approaches~\citep{DeepAR,DeepState,DSSMF}. Such models can share parameters across multiple series while retaining a univariate prediction formulation.

\textbf{Multivariate forecasting.} Multivariate forecasting considers multiple variables and their temporal and cross-variable dependencies~\citep{qiu2026comprehensive}. Existing architectures include attention-based models~\citep{Informer,Triformer,patchtst,iTransformer}, decomposition-based models~\citep{Autoformer,zhou2022fedformer,xPatch}, and compact linear or periodic models~\citep{Zengdlinear,lin2024sparsetsf,lincyclenet}. Cross-variable relationships are also studied through graph and hypergraph representations~\citep{MAGNN,MSHyper,AdaMSHyper}, clustering~\citep{qiu2025duet}, and deformable convolutions~\citep{cheng2026ccd}. These approaches offer different ways to organize temporal and channel information, with common evaluation frameworks supporting systematic comparison~\citep{qiu2024tfb}.

\textbf{Forecasting settings and applications.} Beyond regularly sampled point forecasting, recent studies address irregular observations~\citep{liu2026rethinking,qiu2026bridging,liu2026astgi} and predictive uncertainty through latent-variable, diffusion, and flow-based models~\citep{wu2025k2vae,TimeGrad,TSFlow}. Applications include electricity-price prediction~\citep{EPF}, wind power forecasting with systematic evaluation through WPBench~\citep{zhu2026wpbench}, traffic forecasting with external factors~\citep{williams2001multivariate}, and battery remaining-useful-life prediction~\citep{cheng2026metagnsdformer}.

Forecasting with exogenous variables distinguishes the prediction target from the covariates used as evidence. Future-known covariates provide information over the prediction horizon, motivating correlation-based and conditional generative approaches~\citep{qiu2026dag,GCGNet,cheng2026kite}. XMatch exploits this information by organizing historical exogenous--endogenous pattern correspondences and retrieving relevant endogenous responses.

\subsection{Representation Learning and Robust Modeling for Time Series}
\label{app:representation-robustness}

\textbf{Temporal representations and learning objectives.} Recent studies improve forecasting through selective temporal representations~\citep{wu2025srsnet}, decomposition- and frequency-based learning objectives~\citep{qiu2025DBLoss,wang2025iclrfredf}, and distribution-aware alignment~\citep{wang2026iclrdistdf,hu2026bridging}. DTAF~\citep{lu2026dtaf} addresses non-stationarity through temporal stabilization and frequency differencing. Pretrained forecasting models~\citep{TTM,sundial} and correlation-aware adaptation~\citep{cheng2026cora} further support transferable temporal representations. Beyond numerical prediction, time series question answering is explored through tool-augmented reasoning~\citep{wu2026timeart} and pattern-aware alignment with balanced reasoning in PATRA~\citep{lu2026patra}. These directions emphasize the role of representations, learning objectives, and adaptation across temporal tasks.

\textbf{Missing observations and anomalous behavior.} Robust forecasting studies incomplete observations and varying missing rates~\citep{11002729,yu2025merlin,yang2025revisiting}. Anomaly detection instead distinguishes normal temporal structure from unusual behavior, with advances in unified benchmarking~\citep{qiu2025tab}, temporal-frequency representation learning~\citep{wu2025catch,fang2024temporal}, reconstructive contrast and channel interaction~\citep{hu2024multirc,huteamWork}, and foundation-model adaptation~\citep{STAR}. These studies motivate representations that preserve informative temporal and channel structure under imperfect observations.

\clearpage

\section{Full results}
\label{Full results}

\subsection{Forecasting with Future Exogenous Variables}
\label{app:forecasting_results}

\begin{table}[H]
\centering
\renewcommand{\arraystretch}{1.1}
\caption{Full results on forecasting with historical and future exogenous variables across 12 real-world datasets, where the inputs are ($X^{\text{endo}}, X^{\text{exo}}$, and $Y^{\text{exo}}$). \textcolor{red}{\textbf{Red}}: the best, \textcolor{blue}{\underline{Blue}}: the 2nd best. Avg means the average results from two forecasting horizons.}
\label{tab:main_results_full}
\setlength{\tabcolsep}{4.495pt}
\resizebox{\textwidth}{!}{%
\begin{tabular}{cc|cc|cc|cc|cc|cc|cc|cc|cc|cc|cc|cc}
\toprule
\multicolumn{2}{c}{Models} & \multicolumn{2}{c}{XMatch} & \multicolumn{2}{c}{DAG} & \multicolumn{2}{c}{KITE} & \multicolumn{2}{c}{GCGNet} & \multicolumn{2}{c}{TimeXer} & \multicolumn{2}{c}{TFT} & \multicolumn{2}{c}{TiDE} & \multicolumn{2}{c}{DUET} & \multicolumn{2}{c}{CrossLinear} & \multicolumn{2}{c}{Amplifier} & \multicolumn{2}{c}{TimeKAN} \\
\multicolumn{2}{c}{Metrics} & \multicolumn{1}{c}{mse} & \multicolumn{1}{c}{mae} & \multicolumn{1}{c}{mse} & \multicolumn{1}{c}{mae} & \multicolumn{1}{c}{mse} & \multicolumn{1}{c}{mae} & \multicolumn{1}{c}{mse} & \multicolumn{1}{c}{mae} & \multicolumn{1}{c}{mse} & \multicolumn{1}{c}{mae} & \multicolumn{1}{c}{mse} & \multicolumn{1}{c}{mae} & \multicolumn{1}{c}{mse} & \multicolumn{1}{c}{mae} & \multicolumn{1}{c}{mse} & \multicolumn{1}{c}{mae} & \multicolumn{1}{c}{mse} & \multicolumn{1}{c}{mae} & \multicolumn{1}{c}{mse} & \multicolumn{1}{c}{mae} & \multicolumn{1}{c}{mse} & \multicolumn{1}{c}{mae} \\
\midrule
\multirow[c]{3}{*}{\rotatebox{90}{NP}} & 24 & \textcolor{blue}{\underline{0.185}} & \textcolor{red}{\textbf{0.221}} & 0.202 & 0.237 & \textcolor{red}{\textbf{0.179}} & \textcolor{blue}{\underline{0.223}} & 0.208 & 0.237 & 0.236 & 0.266 & 0.219 & 0.249 & 0.284 & 0.301 & 0.246 & 0.287 & 0.210 & 0.266 & 0.252 & 0.303 & 0.273 & 0.310 \\
 & 360 & \textcolor{red}{\textbf{0.380}} & \textcolor{red}{\textbf{0.377}} & 0.521 & 0.451 & \textcolor{blue}{\underline{0.471}} & \textcolor{blue}{\underline{0.422}} & 0.531 & 0.459 & 0.600 & 0.475 & 0.539 & 0.501 & 0.601 & 0.498 & 0.576 & 0.528 & 0.531 & 0.508 & 0.587 & 0.534 & 0.538 & 0.529 \\
\cmidrule(lr){2-24}
 & Avg & \textcolor{red}{\textbf{0.282}} & \textcolor{red}{\textbf{0.299}} & 0.362 & 0.344 & \textcolor{blue}{\underline{0.325}} & \textcolor{blue}{\underline{0.323}} & 0.370 & 0.348 & 0.418 & 0.371 & 0.379 & 0.375 & 0.443 & 0.400 & 0.411 & 0.408 & 0.371 & 0.387 & 0.420 & 0.418 & 0.405 & 0.419 \\
\midrule
\multirow[c]{3}{*}{\rotatebox{90}{PJM}} & 24 & 0.060 & \textcolor{blue}{\underline{0.147}} & \textcolor{blue}{\underline{0.057}} & \textcolor{red}{\textbf{0.143}} & \textcolor{red}{\textbf{0.056}} & \textcolor{red}{\textbf{0.143}} & 0.060 & 0.150 & 0.075 & 0.166 & 0.095 & 0.195 & 0.106 & 0.214 & 0.072 & 0.166 & 0.088 & 0.191 & 0.096 & 0.208 & 0.115 & 0.244 \\
 & 360 & \textcolor{red}{\textbf{0.110}} & \textcolor{red}{\textbf{0.209}} & 0.130 & 0.218 & 0.135 & \textcolor{blue}{\underline{0.214}} & \textcolor{blue}{\underline{0.129}} & 0.223 & 0.140 & 0.231 & 0.133 & 0.219 & 0.177 & 0.279 & 0.131 & 0.228 & 0.135 & 0.254 & 0.177 & 0.285 & 0.162 & 0.281 \\
\cmidrule(lr){2-24}
 & Avg & \textcolor{red}{\textbf{0.085}} & \textcolor{red}{\textbf{0.178}} & \textcolor{blue}{\underline{0.093}} & 0.180 & 0.096 & \textcolor{blue}{\underline{0.179}} & 0.095 & 0.187 & 0.108 & 0.198 & 0.114 & 0.207 & 0.142 & 0.246 & 0.102 & 0.197 & 0.112 & 0.223 & 0.137 & 0.246 & 0.139 & 0.262 \\
\midrule
\multirow[c]{3}{*}{\rotatebox{90}{BE}} & 24 & 0.361 & 0.241 & 0.361 & \textcolor{red}{\textbf{0.229}} & \textcolor{red}{\textbf{0.348}} & \textcolor{blue}{\underline{0.240}} & \textcolor{blue}{\underline{0.350}} & 0.248 & 0.392 & 0.253 & 0.426 & 0.272 & 0.426 & 0.285 & 0.432 & 0.272 & 0.391 & 0.259 & 0.471 & 0.339 & 0.451 & 0.319 \\
 & 360 & \textcolor{red}{\textbf{0.479}} & \textcolor{red}{\textbf{0.304}} & 0.485 & 0.330 & 0.507 & 0.332 & 0.511 & 0.340 & 0.512 & 0.327 & \textcolor{blue}{\underline{0.482}} & \textcolor{blue}{\underline{0.310}} & 0.571 & 0.364 & 0.597 & 0.436 & 0.568 & 0.416 & 0.646 & 0.487 & 0.645 & 0.495 \\
\cmidrule(lr){2-24}
 & Avg & \textcolor{red}{\textbf{0.420}} & \textcolor{red}{\textbf{0.273}} & \textcolor{blue}{\underline{0.423}} & \textcolor{blue}{\underline{0.280}} & 0.428 & 0.286 & 0.431 & 0.294 & 0.452 & 0.290 & 0.454 & 0.291 & 0.498 & 0.325 & 0.515 & 0.354 & 0.479 & 0.337 & 0.559 & 0.413 & 0.548 & 0.407 \\
\midrule
\multirow[c]{3}{*}{\rotatebox{90}{FR}} & 24 & 0.357 & \textcolor{blue}{\underline{0.174}} & 0.355 & \textcolor{red}{\textbf{0.171}} & \textcolor{red}{\textbf{0.311}} & 0.178 & \textcolor{blue}{\underline{0.347}} & 0.188 & 0.366 & 0.208 & 0.543 & 0.253 & 0.418 & 0.255 & 0.384 & 0.251 & 0.390 & 0.226 & 0.459 & 0.348 & 0.454 & 0.296 \\
 & 360 & \textcolor{blue}{\underline{0.463}} & \textcolor{red}{\textbf{0.258}} & 0.473 & 0.268 & \textcolor{red}{\textbf{0.462}} & 0.271 & 0.482 & 0.279 & 0.489 & 0.273 & 0.465 & \textcolor{blue}{\underline{0.261}} & 0.551 & 0.308 & 0.607 & 0.403 & 0.575 & 0.370 & 0.648 & 0.468 & 0.641 & 0.452 \\
\cmidrule(lr){2-24}
 & Avg & \textcolor{blue}{\underline{0.410}} & \textcolor{red}{\textbf{0.216}} & 0.414 & \textcolor{blue}{\underline{0.219}} & \textcolor{red}{\textbf{0.387}} & 0.225 & 0.415 & 0.234 & 0.427 & 0.241 & 0.504 & 0.257 & 0.484 & 0.281 & 0.496 & 0.327 & 0.483 & 0.298 & 0.554 & 0.408 & 0.547 & 0.374 \\
\midrule
\multirow[c]{3}{*}{\rotatebox{90}{DE}} & 24 & \textcolor{blue}{\underline{0.271}} & \textcolor{red}{\textbf{0.320}} & 0.277 & \textcolor{blue}{\underline{0.322}} & \textcolor{red}{\textbf{0.262}} & 0.327 & 0.280 & 0.331 & 0.339 & 0.362 & 0.380 & 0.383 & 0.367 & 0.383 & 0.376 & 0.378 & 0.387 & 0.396 & 0.394 & 0.407 & 0.399 & 0.412 \\
 & 360 & \textcolor{red}{\textbf{0.427}} & \textcolor{blue}{\underline{0.418}} & 0.462 & \textcolor{blue}{\underline{0.418}} & \textcolor{blue}{\underline{0.437}} & \textcolor{red}{\textbf{0.413}} & 0.523 & 0.447 & 0.610 & 0.474 & 0.599 & 0.509 & 0.630 & 0.511 & 0.589 & 0.482 & 0.583 & 0.507 & 0.551 & 0.474 & 0.547 & 0.479 \\
\cmidrule(lr){2-24}
 & Avg & \textcolor{red}{\textbf{0.349}} & \textcolor{red}{\textbf{0.369}} & 0.370 & \textcolor{blue}{\underline{0.370}} & \textcolor{blue}{\underline{0.350}} & \textcolor{blue}{\underline{0.370}} & 0.401 & 0.389 & 0.475 & 0.418 & 0.489 & 0.446 & 0.499 & 0.447 & 0.482 & 0.430 & 0.485 & 0.452 & 0.473 & 0.441 & 0.473 & 0.445 \\
\midrule
\multirow[c]{3}{*}{\rotatebox{90}{Energy}} & 24 & \textcolor{red}{\textbf{0.061}} & \textcolor{red}{\textbf{0.191}} & 0.079 & 0.215 & \textcolor{blue}{\underline{0.071}} & \textcolor{blue}{\underline{0.208}} & 0.081 & 0.221 & 0.122 & 0.273 & 0.093 & 0.235 & 0.103 & 0.248 & 0.117 & 0.283 & 0.241 & 0.418 & 0.138 & 0.306 & 0.135 & 0.298 \\
 & 360 & \textcolor{red}{\textbf{0.117}} & \textcolor{red}{\textbf{0.271}} & 0.169 & 0.320 & \textcolor{blue}{\underline{0.151}} & \textcolor{blue}{\underline{0.305}} & 0.182 & 0.334 & 0.204 & 0.357 & 0.167 & 0.331 & 0.202 & 0.355 & 0.288 & 0.452 & 0.237 & 0.385 & 0.328 & 0.472 & 0.302 & 0.464 \\
\cmidrule(lr){2-24}
 & Avg & \textcolor{red}{\textbf{0.089}} & \textcolor{red}{\textbf{0.231}} & 0.124 & 0.267 & \textcolor{blue}{\underline{0.111}} & \textcolor{blue}{\underline{0.257}} & 0.131 & 0.277 & 0.163 & 0.315 & 0.130 & 0.283 & 0.153 & 0.302 & 0.203 & 0.367 & 0.239 & 0.402 & 0.233 & 0.389 & 0.218 & 0.381 \\
\midrule
\multirow[c]{3}{*}{\rotatebox{90}{Sdwpfm1}} & 24 & 0.380 & \textcolor{blue}{\underline{0.409}} & \textcolor{red}{\textbf{0.351}} & \textcolor{red}{\textbf{0.400}} & 0.384 & 0.410 & 0.376 & 0.415 & 0.558 & 0.533 & 0.366 & 0.421 & 0.474 & 0.488 & 0.551 & 0.564 & \textcolor{blue}{\underline{0.355}} & 0.473 & 0.364 & 0.445 & 0.418 & 0.503 \\
 & 360 & \textcolor{red}{\textbf{0.423}} & \textcolor{red}{\textbf{0.463}} & 0.495 & 0.522 & \textcolor{blue}{\underline{0.450}} & \textcolor{blue}{\underline{0.491}} & 0.472 & 0.499 & 0.845 & 0.684 & 0.597 & 0.528 & 0.492 & 0.526 & 0.646 & 0.577 & 0.497 & 0.532 & 0.510 & 0.535 & 0.476 & 0.566 \\
\cmidrule(lr){2-24}
 & Avg & \textcolor{red}{\textbf{0.402}} & \textcolor{red}{\textbf{0.436}} & 0.423 & 0.461 & \textcolor{blue}{\underline{0.417}} & \textcolor{blue}{\underline{0.451}} & 0.424 & 0.457 & 0.701 & 0.609 & 0.482 & 0.474 & 0.483 & 0.507 & 0.599 & 0.570 & 0.426 & 0.502 & 0.437 & 0.490 & 0.447 & 0.534 \\
\midrule
\multirow[c]{3}{*}{\rotatebox{90}{Sdwpfm2}} & 24 & 0.446 & 0.453 & \textcolor{red}{\textbf{0.372}} & \textcolor{red}{\textbf{0.414}} & 0.443 & 0.450 & 0.421 & \textcolor{blue}{\underline{0.441}} & 0.627 & 0.570 & 0.411 & 0.458 & 0.461 & 0.492 & 0.445 & 0.452 & 0.477 & 0.536 & \textcolor{blue}{\underline{0.394}} & 0.462 & 0.474 & 0.538 \\
 & 360 & \textcolor{red}{\textbf{0.492}} & \textcolor{red}{\textbf{0.505}} & 0.583 & 0.556 & \textcolor{blue}{\underline{0.506}} & \textcolor{blue}{\underline{0.515}} & 0.529 & 0.531 & 0.978 & 0.736 & 0.541 & 0.519 & 0.511 & 0.540 & 0.584 & 0.528 & 0.589 & 0.611 & 0.587 & 0.563 & 0.520 & 0.589 \\
\cmidrule(lr){2-24}
 & Avg & \textcolor{red}{\textbf{0.469}} & \textcolor{red}{\textbf{0.479}} & 0.477 & 0.485 & \textcolor{blue}{\underline{0.475}} & \textcolor{blue}{\underline{0.483}} & \textcolor{blue}{\underline{0.475}} & 0.486 & 0.803 & 0.653 & 0.476 & 0.488 & 0.486 & 0.516 & 0.514 & 0.490 & 0.533 & 0.573 & 0.491 & 0.512 & 0.497 & 0.564 \\
\midrule
\multirow[c]{3}{*}{\rotatebox{90}{Sdwpfh1}} & 24 & \textcolor{red}{\textbf{0.391}} & \textcolor{red}{\textbf{0.424}} & 0.408 & \textcolor{blue}{\underline{0.438}} & 0.424 & 0.464 & 0.435 & 0.486 & 0.651 & 0.587 & \textcolor{blue}{\underline{0.401}} & 0.460 & 0.434 & 0.489 & 0.527 & 0.513 & 0.548 & 0.585 & 0.576 & 0.627 & 0.511 & 0.582 \\
 & 360 & \textcolor{red}{\textbf{0.422}} & \textcolor{red}{\textbf{0.461}} & 0.489 & 0.534 & \textcolor{blue}{\underline{0.457}} & 0.517 & 0.465 & \textcolor{blue}{\underline{0.514}} & 0.841 & 0.700 & 0.557 & 0.523 & 0.472 & 0.527 & 0.551 & 0.519 & 0.566 & 0.601 & 0.497 & 0.569 & 0.643 & 0.694 \\
\cmidrule(lr){2-24}
 & Avg & \textcolor{red}{\textbf{0.406}} & \textcolor{red}{\textbf{0.442}} & 0.448 & \textcolor{blue}{\underline{0.486}} & \textcolor{blue}{\underline{0.441}} & 0.491 & 0.450 & 0.500 & 0.746 & 0.643 & 0.479 & 0.491 & 0.453 & 0.508 & 0.539 & 0.516 & 0.557 & 0.593 & 0.537 & 0.598 & 0.577 & 0.638 \\
\midrule
\multirow[c]{3}{*}{\rotatebox{90}{Sdwpfh2}} & 24 & \textcolor{red}{\textbf{0.417}} & \textcolor{red}{\textbf{0.446}} & \textcolor{blue}{\underline{0.438}} & \textcolor{blue}{\underline{0.465}} & 0.477 & 0.498 & 0.473 & 0.506 & 0.820 & 0.677 & 0.474 & 0.493 & 0.579 & 0.553 & 0.629 & 0.563 & 0.468 & 0.540 & 0.473 & 0.533 & 0.580 & 0.614 \\
 & 360 & \textcolor{red}{\textbf{0.409}} & \textcolor{red}{\textbf{0.457}} & 0.608 & 0.595 & \textcolor{blue}{\underline{0.524}} & 0.555 & 0.566 & 0.565 & 0.962 & 0.761 & 0.657 & \textcolor{blue}{\underline{0.549}} & 0.619 & 0.614 & 0.665 & 0.569 & 0.608 & 0.609 & 0.569 & 0.628 & 0.713 & 0.729 \\
\cmidrule(lr){2-24}
 & Avg & \textcolor{red}{\textbf{0.413}} & \textcolor{red}{\textbf{0.451}} & 0.523 & 0.530 & \textcolor{blue}{\underline{0.501}} & 0.527 & 0.520 & 0.536 & 0.891 & 0.719 & 0.566 & \textcolor{blue}{\underline{0.521}} & 0.599 & 0.583 & 0.647 & 0.566 & 0.538 & 0.574 & 0.521 & 0.581 & 0.647 & 0.672 \\
\midrule
\multirow[c]{3}{*}{\rotatebox{90}{Colbun}} & 10 & 0.074 & \textcolor{red}{\textbf{0.093}} & \textcolor{blue}{\underline{0.061}} & \textcolor{blue}{\underline{0.094}} & \textcolor{red}{\textbf{0.055}} & 0.101 & 0.065 & 0.108 & 0.113 & 0.172 & 0.092 & 0.135 & 0.089 & 0.131 & 0.089 & 0.134 & 0.071 & 0.102 & 0.071 & 0.121 & \textcolor{blue}{\underline{0.061}} & 0.101 \\
 & 30 & \textcolor{blue}{\underline{0.122}} & \textcolor{red}{\textbf{0.206}} & 0.135 & \textcolor{blue}{\underline{0.215}} & \textcolor{red}{\textbf{0.121}} & 0.243 & 0.149 & 0.243 & 0.176 & 0.299 & 0.383 & 0.460 & 0.240 & 0.322 & 0.307 & 0.397 & 0.182 & 0.288 & 0.275 & 0.370 & 0.195 & 0.249 \\
\cmidrule(lr){2-24}
 & Avg & \textcolor{blue}{\underline{0.098}} & \textcolor{red}{\textbf{0.149}} & \textcolor{blue}{\underline{0.098}} & \textcolor{blue}{\underline{0.154}} & \textcolor{red}{\textbf{0.088}} & 0.172 & 0.107 & 0.175 & 0.145 & 0.235 & 0.238 & 0.297 & 0.164 & 0.227 & 0.198 & 0.266 & 0.126 & 0.195 & 0.173 & 0.246 & 0.128 & 0.175 \\
\midrule
\multirow[c]{3}{*}{\rotatebox{90}{Rapel}} & 10 & 0.187 & 0.213 & \textcolor{red}{\textbf{0.151}} & \textcolor{red}{\textbf{0.203}} & 0.192 & 0.231 & 0.211 & 0.230 & 0.301 & 0.308 & 0.201 & 0.253 & 0.228 & 0.271 & 0.174 & 0.219 & \textcolor{blue}{\underline{0.163}} & \textcolor{blue}{\underline{0.209}} & 0.181 & 0.227 & 0.174 & 0.231 \\
 & 30 & \textcolor{red}{\textbf{0.289}} & \textcolor{red}{\textbf{0.355}} & 0.309 & 0.408 & \textcolor{blue}{\underline{0.296}} & \textcolor{blue}{\underline{0.358}} & 0.401 & 0.384 & 0.387 & 0.416 & 0.409 & 0.414 & 0.411 & 0.432 & 0.365 & 0.432 & 0.340 & 0.417 & 0.333 & 0.416 & 0.325 & 0.390 \\
\cmidrule(lr){2-24}
 & Avg & \textcolor{blue}{\underline{0.238}} & \textcolor{red}{\textbf{0.284}} & \textcolor{red}{\textbf{0.230}} & 0.305 & 0.244 & \textcolor{blue}{\underline{0.295}} & 0.306 & 0.307 & 0.344 & 0.362 & 0.305 & 0.333 & 0.320 & 0.351 & 0.269 & 0.326 & 0.252 & 0.313 & 0.257 & 0.321 & 0.249 & 0.311 \\
\midrule
\multicolumn{2}{c|}{1\textsuperscript{st} Count} & \textcolor{red}{\textbf{22}} & \textcolor{red}{\textbf{29}} & 4 & \textcolor{blue}{\underline{6}} & \textcolor{blue}{\underline{10}} & 2 & 0 & 0 & 0 & 0 & 0 & 0 & 0 & 0 & 0 & 0 & 0 & 0 & 0 & 0 & 0 & 0 \\
\bottomrule
\end{tabular}%
}
\end{table}

\end{document}